\documentclass[a4paper]{cas-dc}

\usepackage[authoryear,longnamesfirst]{natbib}

\def\tsc#1{\csdef{#1}{\textsc{\lowercase{#1}}\xspace}}
\tsc{WGM}
\tsc{QE}

\usepackage{enumitem}
\usepackage{threeparttable}
\usepackage{booktabs}  
\usepackage{multirow} 
\usepackage{makecell}
\usepackage{placeins} 
\usepackage{graphicx}
\usepackage{subcaption}
\usepackage{dsfont}
\usepackage{afterpage}
\usepackage{cuted}
\usepackage{capt-of}

\ExplSyntaxOn
\cs_gset:Npn \__first_footerline:
  {
    \group_begin:
    \small\sffamily
    \__short_authors:
    \group_end:
  }
\ExplSyntaxOff

\begin{document}

\let\WriteBookmarks\relax
\def\floatpagepagefraction{1}
\def\textpagefraction{.001}

\shorttitle{Foundation vs. Task-Specific Models for Probabilistic EPF}    

\shortauthors{Lettner et al.}  

\title [mode = title]{Assessing the Performance--Efficiency Trade-off of Foundation Models in Probabilistic Electricity Price Forecasting}  



%
\author[1]{Jan Niklas Lettner}[orcid=0009-0004-5995-4607]
\cormark[1]
\fnmark[1]
\ead{jan.lettner@kit.edu}
\credit{
Conceptualization,
Formal analysis,
Investigation,
Methodology,
Software,
Validation,
Visualization,
Writing -- original draft,
Writing -- review \& editing
}

\author[1]{Hadeer El Ashhab}[orcid=0009-0000-5132-0504]
\fnmark[1]
\ead{hadeer.elashhab@kit.edu}
\credit{
Conceptualization,
Data curation,
Investigation,
Methodology,
Project administration,
Supervision,
Writing -- original draft,
Writing -- review \& editing
}

\author[1]{Veit Hagenmeyer}[orcid=0000-0002-3572-9083]
\ead{veit.hagenmeyer@kit.edu}
\credit{
Writing -- review \& editing
}

\author[1]{Benjamin Schäfer}[orcid=0000-0003-1607-9748]
\ead{benjamin.schaefer@kit.edu}
\credit{
Conceptualization,
Project administration,
Supervision,
Writing -- review \& editing
}

\affiliation[1]{
    organization={Karlsruhe Institute of Technology (KIT)},
    addressline={Kaiserstraße 12},
    city={Karlsruhe},
    postcode={76131},
    state={Baden-Württemberg},
    country={Germany}
}

\cortext[1]{Corresponding author}
\fntext[1]{These authors contributed equally to this work.}

\begin{abstract}
Large-scale renewable deployment adds volatility to electricity prices. Probabilistic forecasts that quantify this uncertainty support trading, dispatch, and risk management, but it remains unclear which models produce the best forecasts for probabilistic electricity price forecasting (PEPF). Time series foundation models (TSFMs) promise strong performance without task-specific training, whereas task-specific deep learning models remain standard. We compare 4 models for day-ahead PEPF in European bidding zones: NHITS with quantile regression averaging (NHITS+QRA), a Transformer-conditioned normalizing flow, and 2 TSFMs, namely Moirai and ChronosX. Models are compared using proper scoring rules, namely the continuous ranked probability score and the energy score, together with calibration diagnostics. When fine-tuned on prices and calendar features alone, the TSFMs attain the best accuracy and calibration. However, a well-configured NHITS+QRA closes most of this gap with roughly 3 orders of magnitude fewer parameters. It is the only model that converts additional exogenous feature groups, such as load, generation mix, and cross-border flows, into improved accuracy, whereas Moirai degrades markedly when given the same groups. When models are instead pretrained on 14 neighboring bidding zones and adapted to DE-LU with limited target data, NHITS+QRA reaches parity with ChronosX, consumes roughly 1 order of magnitude less training energy than Moirai, and attains the lowest expected calibration error. The ranking of TSFMs and task-specific models therefore depends on the available features and the adaptation regime. As our comparison rests on single-seed runs on one target zone, we report these rankings as indicative. We recommend weighing computational cost against marginal gains when selecting a PEPF model.
\end{abstract}




\maketitle

\section{Introduction}
The accelerating pace of climate change has placed renewable energy at the forefront of global mitigation strategies. Nations worldwide are committing to large-scale deployment of wind, solar, and other clean power sources, with the common goal of decarbonizing the electricity sector. This transition, however, brings new technical challenges: renewable generation is intrinsically variable, producing power that fluctuates across minutes, hours, and seasons. The resulting volatility propagates through the power system, creating substantial uncertainty for market participants, grid operators, and policy makers alike~\cite{putzFeasibilityForecastingHighly2024,cramerMultivariateProbabilisticForecasting2023b,nationsRenewableEnergyPoweringa}.

Reliable forecasts, especially probabilistic forecasts that capture the full distribution of future outcomes, have therefore become indispensable tools for managing risk, scheduling resources, and ensuring system stability in high-renewable contexts~\cite{cramerMultivariateProbabilisticForecasting2023b,trebbienProbabilisticForecastingDayAhead2023}.

Among the most immediately impactful applications of such forecasts is electricity price forecasting (EPF). In markets with growing shares of stochastic renewable output, spot and intraday prices are driven not only by demand and fuel costs but also by uncertainty on the supply side. This makes probabilistic price forecasts economically valuable: they support hedging, bidding, and contracting decisions for generators, retailers, and traders by quantifying risk rather than providing a single-point estimate~\cite{AliyonRitvanen2024}. Beyond market profits and risk management, prices also act as a coordination signal for system operation. Short-term stability is maintained by fast, rule-based control, but medium- to longer-term balancing and flexibility are increasingly procured through markets. Scarcity and imbalance conditions therefore propagate into prices that incentivize corrective actions. Robust probabilistic EPF (PEPF) is therefore central to both efficient market participation and reliable system balancing~\cite{trebbienUnderstandingElectricityPrices2023a,AliyonRitvanen2024}.

Europe’s integrated wholesale electricity market provides a \allowbreak uniquely demanding testbed for PEPF. Day-ahead prices are formed across multiple interconnected bidding zones via market coupling. Forecasts must therefore remain robust under cross-border exchanges, transmission constraints, and heterogeneous renewable portfolios~\cite{cramerMultivariateProbabilisticForecasting2023b,ElectricityMarketDesign}. At the same time, Europe offers unusually rich public data--e.g., the ENTSO-E transparency framework--enabling systematic enrichment of autoregressive price models with fundamentals such as generation, load, and cross-border flows~\cite{MissionStatement}. Within this setting, we fix Germany-Luxembourg (DE-LU) as our test zone. As a central and highly active market area, DE-LU provides a realistic, high-impact environment for evaluating feature-driven probabilistic forecasts under large-scale renewable integration. Importantly, DE-LU sits at the core of Germany's energy transition (Energiewende). Here, rapid renewable expansion and ongoing system transformation make price dynamics particularly sensitive to policy and structural change---exactly the conditions under which robust probabilistic forecasting is most valuable~\cite{BiddingZoneReviewa}.

In recent years, the ML community has increasingly shifted from specialized models, trained separately for each dataset and task, toward foundation models. Foundation models are large neural architectures pre-trained on diverse, large-scale data collections; they learn transferable representations that can be reused across many downstream applications~\cite{StanfordCRFM}.
This paradigm, already well established in natural language processing and computer vision~\cite{paassFoundationModelsNatural2023,FoundationModelsDefining}, reduces the need for heavy task-specific feature engineering and enables strong performance even when labeled data or domain-specific training budgets are limited. For a given downstream problem, foundation models can be applied directly without additional training (zero-shot) or fine-tuned more extensively when sufficient data is available. In time-series forecasting, recent foundation models such as Moirai \cite{wooUnifiedTrainingUniversal2024} and ChronosX \cite{arangoChronosXAdaptingPretrained2025} follow this approach. They pre-train on broad time-series corpora to capture generic temporal patterns such as seasonality, regime changes, and non-linear dependencies. This makes them attractive candidates for PEPF, where the target dynamics are complex and shift across markets and time. Their transfer-learning capabilities also enable a systematic comparison of zero-shot, few-shot, and full fine-tuning regimes in electricity markets. Such a comparison helps quantify how much task-specific adaptation is needed beyond general time-series pretraining.

The present work seeks to benchmark these time series foundation models (TSFMs) against state-of-the-art (SOTA) probabilistic forecasting techniques that are exclusively trained on this dataset. Specifically, we consider: 1) NHITS+QRA, a hybrid probabilistic forecasting approach that combines NHITS (Neural Hierarchical Interpolation for Time Series)--a deep learning forecaster with a Multi-rate sampling with hierarchical interpolation ~\cite{challuNHITSNeuralHierarchical2023a}--with Quantile Regression Averaging (QRA), a post-processing technique used to construct calibrated predictive intervals from point forecasts~\cite{nowotarskiComputingElectricitySpot2015}. 2) normalizing flow (NF)~\cite{papamakariosNormalizingFlowsProbabilistic2021} which is a deep generative modeling framework that learns a bijective transformation from a simple base distribution to the conditional distribution of electricity prices, thereby delivering flexible, high-resolution density forecasts.

By comparing Moirai and ChronosX against these SOTA models, we can assess whether the expressive power of TSFMs translates into measurable gains in forecast accuracy, calibration and computational efficiency for PEPF. We believe that our work addresses several gaps in the literature. First, while prior studies have mainly evaluated TSFMs in terms of point forecasts, we assess their probabilistic forecasting performance in electricity markets, an angle that has received comparatively little attention. Second, we compare different learning strategies, including zero-shot, one-shot, and few-shot adaptations, to understand how transfer from other regions or markets affects forecast quality under realistic information constraints. Third, we explore NHITS combined with QRA as a strong, computationally efficient baseline, which, to our knowledge, has not been applied in this context before. Finally, we consider the potential of a Transformer-conditioned NF trained by minimizing the energy score, rather than relying solely on likelihood-based training, to produce well-calibrated predictive distributions.

The rest of the paper is organized as follows: Section~\ref{sec:rel} reviews related work, while Section~\ref{sec:data} describes the dataset. Section~\ref{sec:frame} details the four implemented forecasting models. Section~\ref{sec:exp} and Section~\ref{sec:res} outline the experimental setup and analyze performance. Finally, Section~\ref{sec:sum} concludes with implications for renewable-inclusive power systems.

\begin{figure*}
  \includegraphics[width=\textwidth]{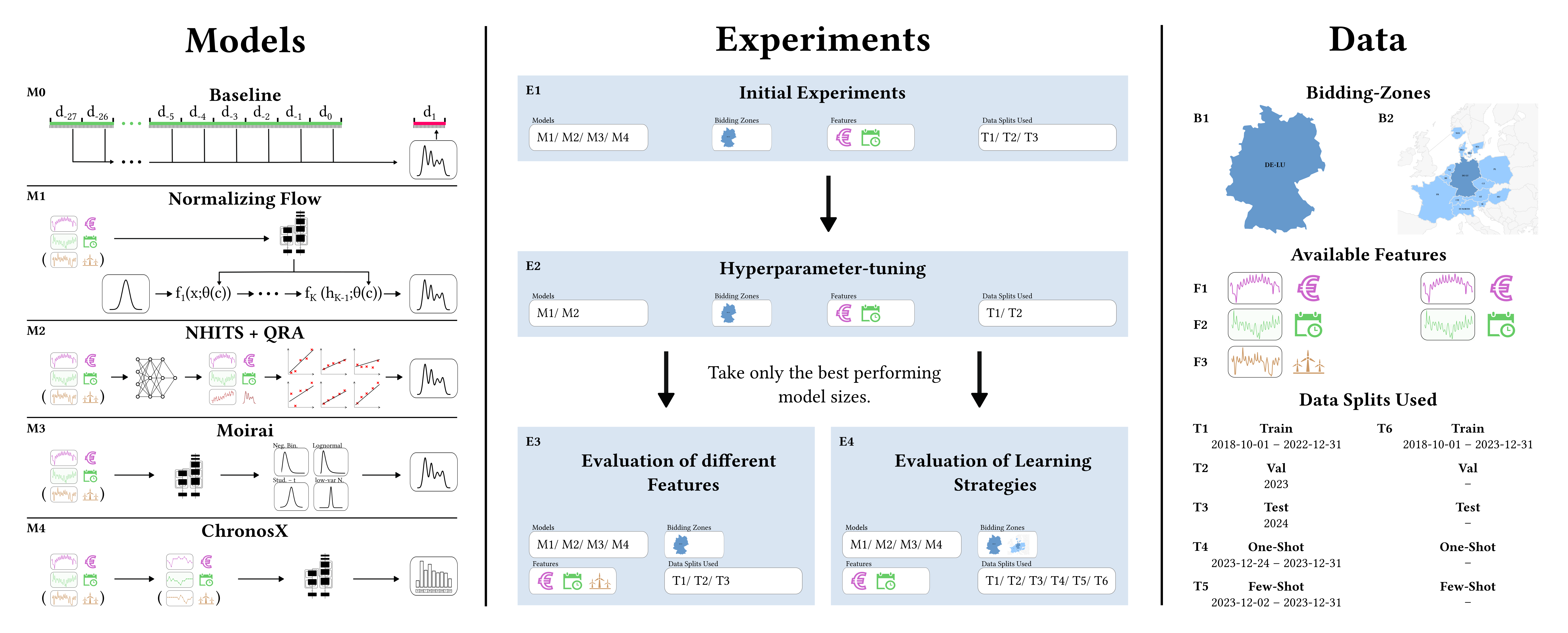}
    \caption{Graphical Abstract. The left column lists the evaluated models: a simple baseline (M0), a conditioned NF (M1), NHITS with a QRA head (M2), and two foundation models--Moirai (M3) and ChronosX (M4). The right column summarizes the datasets. We use either only the DE-LU bidding zone (B1) or DE-LU plus 14 neighboring bidding zones (B2). For B1, three feature groups are available: electricity prices (F1), calendar features (F2), and past-only fundamentals such as load and renewable share (F3). The F3 feature group is not available for the additional bidding zones in B2. The dataset splits are also shown: DE-LU is divided into training (T1), validation (T2), and test (T3) sets; to study varying exposure to DE-LU data, we further define one-shot (T4) and few-shot (T5) settings. Data from the other bidding zones are denoted by T6. The middle column depicts four experiment blocks. E1 establishes a baseline by training or fine-tuning each model on DE-LU with calendar features only, evaluating three model sizes per family (tiny/small/base for the proposed models; small/base/large for the foundation models). In E2, we tune the hyperparameters of the proposed models and select the best configuration per size. Using these fixed configurations, E3 examines how different feature groups affect forecasting performance, and E4 how pretraining on neighboring bidding zones and varying DE-LU exposure (zero-shot/one-shot/few-shot) influence results.
    }
  \label{fig:big}
\end{figure*}

\section{Related Work}
\label{sec:rel}

EPF has evolved from early statistical-machine-learning hybrids~\cite{andradeProbabilisticPriceForecasting2017} to deep-learning architectures capable of capturing non-linear market dynamics. Large-scale studies indicate that deep-learning--based forecasting substantially alters traditional predictability patterns across European bidding zones~\cite{AliyonRitvanen2024}. Recurrent architectures such as LSTMs effectively model price levels and volatility during market stress~\cite{trebbienProbabilisticForecastingDayAhead2023}, while distributional neural networks and NF further enhance uncertainty modeling by learning full predictive distributions~\cite{marcjaszDistributionalNeuralNetworks2023a, cramerMultivariateProbabilisticForecasting2023b}.

A foundational reference for rigorous EPF benchmarking compares SOTA statistical and deep-learning methods across multiple years and markets, proposes a set of best practices, and releases an open-access benchmark including the LEAR (Lasso Estimated AutoRegressive) and DNN-EPF reference models~\cite{lagoForecastingDayaheadElectricity2021a}. Their work highlights several recurrent shortcomings of the EPF literature for example: short test samples, single-market evaluation, weak baselines, and absence of significance testing. 
We address the first of these with a full year of hourly out-of-sample data (2024), and the last partially through Diebold--Mariano testing during feature selection. We do not fully resolve the remaining two: evaluation is confined to the DE-LU test zone (other zones enter only as pretraining data in E4), and we benchmark against four transparent statistical baselines rather than against LEAR or DNN-EPF. We return to both points in Section~\ref{sec:limitations}.

In parallel, PEPF continues to rely on regression-based post-processing. Quantile regression remains a stable and competitive approach \cite{osoneQuantileRegressionProbabilistic2025}, with \allowbreak{} smoothing extensions providing increased reliability and economic value in trading applications \cite{uniejewskiSmoothingQuantileRegression2025}. QRA \allowbreak{} \cite{nowotarskiComputingElectricitySpot2015} has emerged as a particularly effective post-processing technique, and recent work has shown that combining strong point forecasters with explicit quantile calibration heads yields highly competitive PEPF performance~\cite{marcjaszDistributionalNeuralNetworks2023a}. Probabilistic forecasting also matters beyond short-term trading: it informs medium- and long-term policy and investment decisions~\cite{zielProbabilisticMidLongterm2018}.

Recent research has pivoted toward cross-market generalization and interpretability. Transfer learning via pretraining and fine-tuning has been shown to improve accuracy across similar European day-ahead markets~\cite{gunduzTransferLearningElectricity2023a}, and hybrid models that integrate deep learning with sparse feature selection demonstrate improved stability and robustness~\cite{jiangProbabilisticElectricityPrice2025}.

In the broader time-series community, the rise of pretrained foundation models has prompted renewed scrutiny of whether large, language-model-style architectures actually outperform simpler approaches~\cite{NEURIPS2024_6ed5bf44}. Those authors report that, on standard time-series benchmarks, language-model-based forecasters often fail to provide measurable benefits over much simpler baselines, casting doubt on the assumption that scale and architectural sophistication automatically translate into better forecasts. To our knowledge, this question has not yet been examined in the specific context of PEPF.

Our work addresses this gap by providing a unified probabilistic evaluation that compares two recent TSFMs (Moirai and ChronosX) against carefully tuned task-specific models (NF and NHITS+QRA) under realistic information constraints, on the DE-LU market, and across zero-shot, one-shot, few-shot, and full fine-tuning regimes. 

\section{Data}
\label{sec:data}
\begin{figure}[pos=h]
    \centering
    \includegraphics[width=0.8\linewidth]{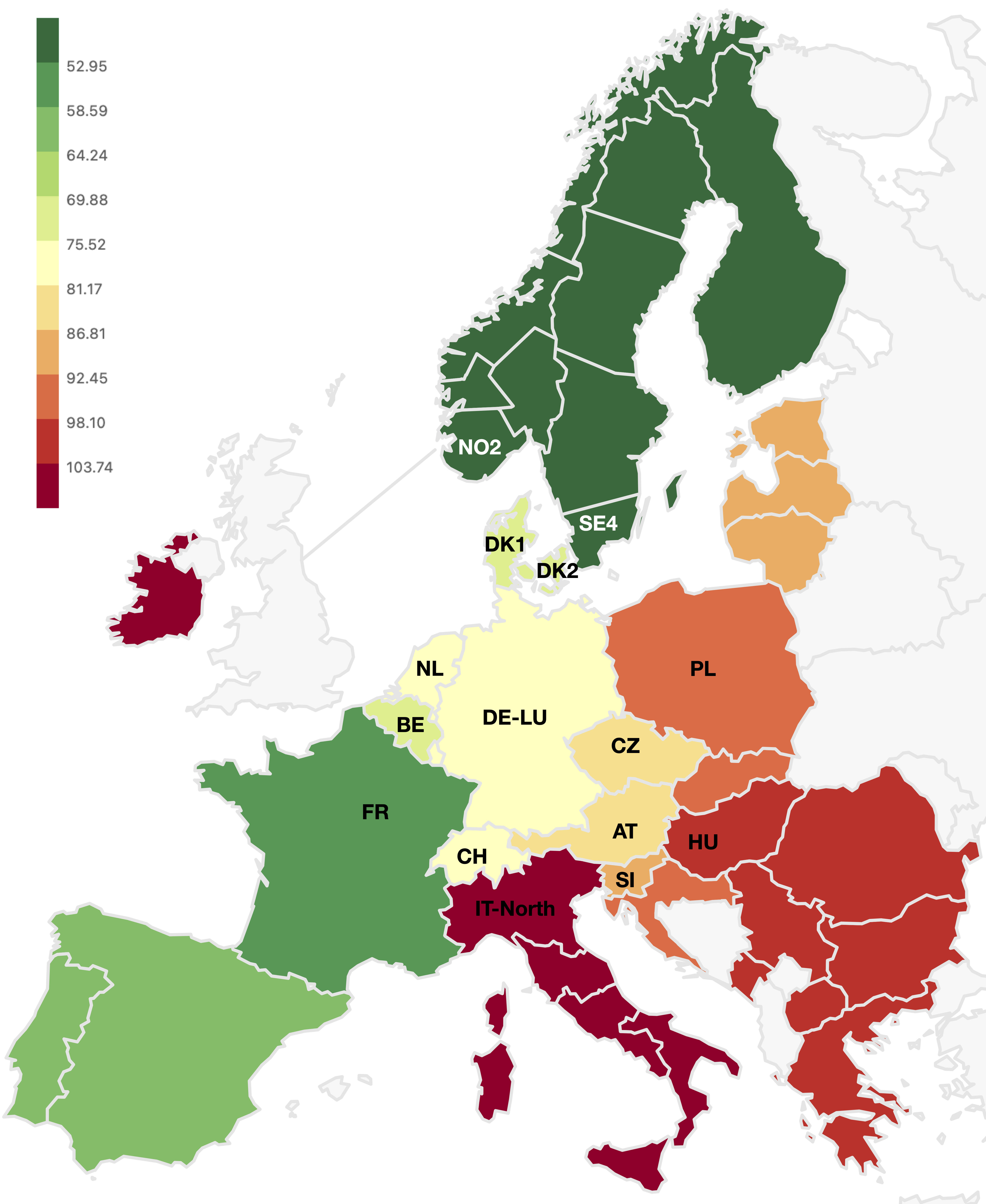}
    \caption{Average electricity spot market prices across European bidding zones in 2024, measured in EUR/MWh. Countries or zones are color-coded based on their average annual prices, with darker red indicating higher prices and darker green indicating lower prices. The labeled bidding zones (e.g., DE-LU, FR, DK1, etc.) were considered in our learning strategies' experiments. Adapted from ~\cite{map_energycharts} with permission.}
    \label{fig:bidding-zones-map}
\end{figure}

In the European wholesale electricity market a bidding zone is a geographically defined area for which a single market-clearing price is determined. In most cases the borders of a bidding zone coincide with national frontiers, but notable exceptions exist: Italy is split into multiple separate zones, while Germany and Luxembourg share a common zone (DE-LU). Figure~\ref{fig:bidding-zones-map} provides a visual overview of the zones considered in our study. Formally, a bidding zone can be described as a geographical area within the electricity market where electricity can be bought and sold without considering physical grid limitations.

We study the DE-LU bidding zone in detail for two main reasons. First, it is the largest bidding zone in Europe, encompassing all of Germany and serving a population of over 80 million people. Second, it is highly interconnected with several other bidding zones, including both continental European and Nordic zones~\cite{trebbienPatternsCorrelationsEuropean2024a,BiddingZoneReviewa}.

The choice of 2024 allows us to examine model behavior under current and dynamic market conditions, characterized by price volatility, policy shifts, and renewable integration. Each quarter includes different types of challenges--ranging from price spikes to periods of relative stability--offering a diverse testbed for assessing forecasting robustness and generalization.

The electricity price data employed in this work are retrieved from~\cite{burgerEnergyChartsa} and its public API~\cite{EnergyChartsAPIa}, both operated by the Fraunhofer Institute for Solar Energy Systems (Fraunhofer ISE). The platform supplies hourly day-ahead spot-market prices for all European bidding zones in \(\mathrm{EUR/MWh}\) with a homogeneous coverage that starts in 2015. Licensing and usage terms are provided in the Data Licensing and Terms of Use statement.

The training set spans from October 2018 to the end of 2022, covering relatively stable periods, the COVID-19 shock, and the 2022 energy crisis~\cite{goldthauEnergyCrisisFive2022a,oziliGlobalEnergyCrisis2023a,trebbienPatternsCorrelationsEuropean2024a}.

All series are sampled at an hourly resolution and prices are expressed in \(\mathrm{EUR/MWh}\). Following the findings of our preceding work \cite{11664416}, the input horizon is fixed to one week (\(24\times7=168\) time steps), which yielded the best performance compared to shorter or longer histories. The forecast horizon is always one day (\(24\) time steps). Consequently, each training/validation/test sample consists of a matrix \(\mathbf{X}\in\mathbb{R}^{168\times F}\) (input) and a vector \(\mathbf{y}\in\mathbb{R}^{24}\) (target), where \(F\) denotes the number of exogenous features described below:

\begin{itemize}
    \item \textbf{Calendar features:} Hour, day of week, and month (cyclical), plus isWeekend and isHoliday (one-hot).
    \item \textbf{Market features:} Gas price, $\mathrm{CO}_2$ allowance price, and synthetic price (see Section~\ref{sec:SyntheticPrice}), and electricity load (MW). All prices are in EUR/MWh.
    \item \textbf{Supply \& trading:} Renewable, non-renewable generation, and net cross-border flows (all in MW).
\end{itemize}


The day-ahead auction on EPEX SPOT closes at 12:00 CET on day $X-1$. To conservatively reflect the information set available before gate closure, all market-, supply-, and trading-related exogenous variables are masked after 10:00 on day $X-1$, providing a two-hour buffer that accounts for the staggered publication and aggregation of fundamentals across the bidding zone. Deterministic calendar attributes remain unmasked for both past and future timestamps.

Following the DE-LU-AT zone redefinition, the dataset spans 2018-10-01 to 2024-12-31, ensuring a clean separation for the DE-LU test zone. The data is partitioned into Training (2018-10-01--2022-12-31), Validation (2023), and Test (2024) sets. For learning-strategy experiments, models are pretrained on all other available bidding zones (2018--2023) and evaluated using varying DE-LU increments: \emph{zero-shot} (no DE-LU samples), \emph{one-shot} (one 8-day/192-hour sample), or \emph{few-shot} (30 days/23 samples from late 2023).

Samples are generated using a sliding window. The training set employs a 1-hour stride to maximize instances via overlapping windows. Conversely, validation and test sets use a 24-hour stride to ensure independent daily forecasts, with the exception of the ChronosX model, which uses a 1-hour stride for validation due to implementation constraints. To ensure an unbiased evaluation, all features and targets are standardized using the mean and standard deviation of the training split, which are then applied to the validation and test sets to prevent information leakage.
\subsection{Problem Formulation}
\label{sec:problem}
We formalize the day-ahead PEPF task as follows. Let $y_t \in \mathbb{R}$ denote the day-ahead electricity spot price in the DE-LU bidding zone at hour $t$ (in EUR/MWh), and let $\mathbf{x}_t \in \mathbb{R}^{F}$ denote the vector of $F$ exogenous covariates available at hour $t$ (calendar features, market features, supply and trading variables, as defined in Section~\ref{sec:data}). For a forecast origin $t_0$, we define the input window
\[
\mathbf{X}_{t_0} = \bigl[(y_{t_0-167}, \mathbf{x}_{t_0-167}),\ \ldots,\ (y_{t_0}, \mathbf{x}_{t_0})\bigr] \in \mathbb{R}^{168 \times (F+1)},
\]
covering the most recent week of hourly observations, and the target vector
\[
\mathbf{y}_{t_0} = (y_{t_0+1}, \ldots, y_{t_0+24}) \in \mathbb{R}^{24},
\]
covering the next $H=24$ hours.

To reflect the realistic information set available at the day-ahead market gate closure, all market-, supply-, and trading-related covariates in $\mathbf{X}_{t_0}$ and over the forecast horizon are masked beyond 10:00 on the forecast origin day; deterministic calendar features remain unmasked. We denote this masking operator by $\mathcal{M}$ and the resulting masked input by $\mathcal{M}(\mathbf{X}_{t_0})$.

The objective is to estimate the conditional predictive distribution
\[
p\bigl(\mathbf{y}_{t_0} \,\big|\, \mathcal{M}(\mathbf{X}_{t_0})\bigr)
\]
over the full $H$-hour day-ahead price trajectory. All models considered in this work, although different in their architecture and training procedure, are evaluated on this same task using the strictly proper scoring rules: Continuous Ranked Probability Score (CRPS) and Energy Score (ES), as well as calibration diagnostics based on Probability Integral Transform (PIT) histograms and Expected Calibration Error (ECE) (see Appendix~\ref{sec:loss_fun_scores}).

\section{Methods}
\label{sec:frame}


This section describes the four forecasting models compared in this study, which fall into two categories. The first category comprises task-specific deep learning models trained from scratch on the target dataset: NF and NHITS+QRA. The second category comprises two pretrained TSFMs, Moirai and ChronosX, which we evaluate both zero-shot and after fine-tuning on the target dataset. We first present the task-specific models, followed by the TSFMs. Simple statistical baselines used as reference points throughout the paper are presented in Appendix~\ref{sec:baselines}.
\subsection{ML Methods}

\paragraph{normalizing flow}

The NF forecaster couples a Transformer encoder-decoder with a conditional masked-auto\-regressive flow (MAF) \cite{papamakariosNormalizingFlowsProbabilistic2021}. 
\begin{itemize}
    \item \textbf{Transformer conditioner.} Past targets $y_{(t-C):t}$ and all covariates are embedded and processed by a Transformer. The decoder receives an optional injection of the last $K$ realized prices and produces a single context vector $\mathbf{c}\in\mathbb{R}^{d}$ that summarizes the whole horizon.
    \item \textbf{Conditional flow head.} $\mathbf{c}$ conditions a stack of $L$ MAF layers (implemented in the normflows package~\cite{stimperNormflowsPyTorchPackage2023}). Each layer consists of a masked affine autoregressive transformation, followed by an ActNorm and a permutation. The base distribution is a trainable $M$-component diagonal Gaussian mixture over the $H=24$ forecasted prices.
\end{itemize}
The overall mapping
\[
\mathbf{z}\sim p_{\theta}(\mathbf{z})\;\;\xrightarrow{\;f^{-1}_{\theta}(\cdot;\mathbf{c})\;}\;\;
\mathbf{\hat{y}}_{1:H}
\]
yields a full joint density for the day-ahead price trajectory.

To reflect the day-ahead market, all exogenous covariates are masked for the 14-hour window \([10{:}00,24{:}00]\) on the forecast origin day; only calendar features (hour, weekday, month, weekend and holiday flags) are available for both past and future timestamps. This causal masking is enforced both in the Transformer encoder (unknown covariates are hidden) and in the decoder attention mask (autoregressive within the prediction block).

For the probabilistic head we employ a conditional MAF. Compared to coupling-layer flows such as RealNVP, autoregressive flows are generally more expressive: each dimension can be transformed using a neural network conditioned on all preceding dimensions under a chosen ordering~\cite{papamakariosNormalizingFlowsProbabilistic2021}. A potential concern is that the autoregressive factorization introduces an ordering in the density representation:
\[
p(\mathbf{y}\mid\mathbf{c}) \;=\; \prod_{h=1}^{H} p(y_h \mid y_{<h}, \mathbf{c}).
\]
Under this factorization, the conditional for $y_h$ is parameterized without explicit access to $y_{>h}$. Different orderings are possible, and the factorization does not imply sequential revelation of prices in real time. Rather, it provides a flexible way to model the day-ahead price vector while respecting the market's information set through the causal covariate masking described above.
Although NF models admit a negative log-likelihood (NLL) loss, in our preliminary experiments minimizing the ES yielded better-calibrated forecasts; we therefore adopt the ES as the training objective. During training, we draw $S$ samples from the conditional flow and estimate the multivariate ES using a projection-based Monte Carlo representation motivated by the Cramér–Wold theorem. Specifically, the predicted and observed $H$-dimensional trajectories are projected onto randomly sampled one-dimensional directions. The corresponding univariate CRPS values are then evaluated and averaged across projection directions and the batch to obtain an estimate of the multivariate energy score.


The NF forecaster learns a flexible, non-Gaussian joint distribution over the 24-hour day-ahead price vector while respecting the realistic information set imposed by the market’s timing constraints. By conditioning a powerful MAF on a Transformer-derived context, the model captures both short-term recency and long-term seasonal patterns. The ES-based training objective further targets good probabilistic calibration for downstream decision-making.

\paragraph{NHITS+QRA}
A deterministic NHITS forecaster produces a 24-hour day-ahead point forecast in a single forward pass. To obtain a full predictive distribution we post-process the NHITS predictions with QRA: for each horizon, we fit LASSO quantile regressions on a design matrix that combines (i) an ensemble of stochastic NHITS forecasts generated either via Monte-Carlo dropout or Stochastic Weight Averaging Gaussian (SWAG)~\cite{maddoxSimpleBaselineBayesian2019a}, (ii) future-known covariates, and (iii) optional static features. When the ensemble dimension becomes large, we optionally apply incremental Principal Component Analysis (PCA) to the stochastic forecast features prior to the quantile fits. The NHITS backbone retains its original Mean Absolute Error (MAE) training objective; distributional calibration is handled entirely by the QRA head.

We obtain the stochastic NHITS ensemble either by enabling Monte-Carlo dropout at inference or by applying SWAG. In the SWAG case, starting from a configurable epoch, we collect Stochastic Gradient Descent (SGD) iterates to form an SWA mean $\theta_{\mathrm{SWA}}$. SWAG then approximates a Gaussian distribution over weights $q(\theta)\approx\mathcal{N}(\theta_{\mathrm{SWA}},\Sigma)$, from which we sample multiple weight realizations to generate the $S$ stochastic point forecasts used by QRA.

The \texttt{NHITS+QRA} data module supplies, for each rolling origin:
\begin{itemize}[noitemsep]
    \item \texttt{y\_past} ($C$) and \texttt{y\_future} ($H$)
    \item past covariates \texttt{c\_ctx\_future\_known} ($C\times d_k$) and \\
          \texttt{c\_ctx\_future\_unknown} ($C\times d_u$)
    \item future-known covariates aligned with the horizon \\
          \texttt{c\_fct\_future\_known} ($H\times d_k$)
    \item optional static vector (\texttt{static}).
\end{itemize}

\begin{enumerate}[noitemsep]
    \item \textbf{NHITS backbone.} Multi-resolution interpolation \allowbreak{} blocks ingest the four tensors above and output the $H$-step point forecast. Training uses MAE loss, AdamW optimizer and a warm-up + cosine LR schedule. At validation we generate an ensemble of $S$ stochastic forecasts per origin.
    \item \textbf{QRA head.} For each horizon $h$ and target quantile level $\tau$, we fit a separate linear quantile regression (one model per $(h,\tau)$) on a design matrix $X_h\in\mathbb{R}^{N_h\times F_h}$ that concatenates the $S$ stochastic draws and horizon-specific/static features (optionally the sample mean/SD). The parameters are estimated by minimizing the pinball loss with an $\ell_1$ penalty (LASSO),
          \[
          \min_{\beta,b}\;\frac{1}{N_h}\sum_{i}\rho_\tau\!\big(
          y_i - X_{h,i}\beta - b\big)+\lambda\|\beta\|_1,
          \]
          yielding calibrated quantiles $\hat q_{h,\tau}$.
\end{enumerate}
Monotonicity across quantile levels is enforced by a row-wise isotonic regression post-process. Each of the quantiles is then linearly interpolated.

The pipeline proceeds as follows: (i) NHITS is trained on the training set with early stopping on validation MAE; (ii) the best checkpoint is reloaded and used to collect $S$ stochastic forecasts for the validation origins; (iii) the QRA design matrices are constructed (optionally reduced by incremental PCA) and the LASSO quantile regressions are fitted; and (iv) the resulting quantiles are evaluated using CRPS, ES (Appendix~\ref{sec:loss_fun_scores}), and PIT histograms.

All steps respect the realistic information set (known/\allowbreak{}unknown covariate masking and the 14-hour day-ahead-market mask) ensuring an unbiased comparison with other probabilistic models.
\subsection{TSFMs}

\paragraph{Moirai}

Moirai is a universal probabilistic forecaster built around an encoder-only Transformer. The model is released in three sizes--\allowbreak small ($\approx$ 14M), base ($\approx$ 91M), and large ($\approx$ 311M)--and is designed to operate across time series with very different sampling frequencies. To support multiple sampling frequencies, Moirai learns distinct input/output projection layers for several patch sizes; the patch size used for a given time-series frequency is selected via pre-defined settings. This keeps attention costs manageable for high-frequency signals while preserving sufficient resolution for lower-frequency series.

A key architectural choice is that Moirai represents the target series and all covariate series as a single flattened token sequence. Individual variables are distinguished by a variate identifier, allowing the Transformer to attend jointly over time and across variables. In effect, the model can learn dependencies not only along the temporal dimension, but also between different covariates and the target through cross-variate attention.

For the probabilistic output layer, Moirai employs a mixture over parametric likelihood families tailored to common data regimes. The mixture combines a Student-$t$ component for robustness, a negative binomial for positive count data, a log-normal for right-skewed quantities (frequent in economic and natural processes), and a low-variance normal component for high-confidence predictions. Training minimizes the (mixture) negative log-likelihood, yielding calibrated predictive distributions without requiring sampling-based heads.

Moirai is pretrained on a very large multi-domain corpus spanning nine application areas (e.g., energy, transport, climate, web, sales, and healthcare), totaling on the order of $2.8\times 10^{10}$ observations. Notably, this corpus does not include dedicated electricity price datasets. The small model is trained for 100k optimization steps, while base and large are trained for 1M steps. Evaluation distinguishes two regimes. In-distribution performance is assessed using dataset-specific train/test splits, while out-of-distribution behavior is assessed by completely excluding selected time series during training, enabling a zero-shot transfer evaluation.

During fine-tuning and inference, we mask market-related covariates over the final 14 hours [10{:}00,24{:}00] of the forecast-origin day and, following Moirai authors' recommended settings for hourly series, use a patch size of 32.

\paragraph{ChronosX}

Chronos reframes time-series forecasting as a language-modeling problem. The observed series is mean-scaled and quantized into discrete bins, yielding a sequence of integer tokens; a standard Transformer language model (the authors of Chronos primarily use the encoder--decoder T5) is then trained with cross-entropy loss to predict the next tokens. In effect, regression is performed via classification. At inference time, Chronos generates probabilistic forecasts by autoregressive sampling of future tokens and mapping them back to numerical values. Repeated sampling yields a predictive distribution over future trajectories.

The released Chronos family spans four sizes: mini ($\approx$20M), small ($\approx$46M), base ($\approx$200M), and large ($\approx$710M). It was pretrained on a broad, multi-domain collection of 55 public datasets covering energy, transport, healthcare, retail, web, weather, and finance, among others. The corpus is partitioned into pretraining-only (13 datasets), in-domain (15), and zero-shot (27) subsets for evaluation.

Notably for EPF, the only pretraining dataset that explicitly includes electricity prices is the Spanish Energy and Weather dataset, which contains Spain-wide prices alongside demand, generation, and weather. The remaining energy datasets primarily cover load, production, solar, and wind time series. The authors of ChronosX also augment scarce real-world data with mixup-style and synthetic generation techniques (e.g., Gaussian-process-based synthesis) to improve generalization. Training is reported on an AWS instance with 8$\times$ A100 (40\,GB) GPUs; training times range from roughly 7.7\,h for mini/small up to 63.1\,h for large.

A practical limitation of vanilla Chronos is that it does not natively consume rich exogenous covariates. In our work we therefore use ChronosX, an add-on that incorporates past-observed and future-known covariates via lightweight injection blocks while keeping the pretrained Chronos backbone largely untouched. This lets us integrate calendar and market-available drivers under the same realistic information constraints as the other models, enabling a fair comparison in the day-ahead setting. As done for Moirai, the 14-hour window [10{:}00,24{:}00] is masked for market-related features.

\section{Experimental Setup}
\label{sec:exp}

We structure our study into four stages. Each stage isolates a specific question (model comparison, hyperparameter optimization, feature importance, and learning strategy), while using the same data splits and evaluation protocol described in Section~\ref{sec:data}.

\subsection{E1: Baseline Comparison with Price-Only Inputs}
\label{sec:comparison}

In the first set of experiments, we compare all forecasting models against the best classical baseline identified in Section~\ref{sec:baselines} (\emph{Same-Hour (Last 28 Days)}). 
The following four methods are evaluated:

\begin{itemize}[leftmargin=1.5em]
    \item \textbf{TSFMs}: Moirai and ChronosX, each instantiated in three fixed variants (small, base, large), which are not further tuned. Each of these six variations (three sizes per TSFM) is evaluated in two modes: zero-shot and fine-tuned. The zero-shot mode uses the TSFM out of the box, with no additional training, while the fine-tuned mode adapts the model to the DE-LU training set described in Section~\ref{sec:data}.
    \item \textbf{Deep learning models}: NF and NHITS\allowbreak+QRA, each initially evaluated using default configurations. These parameters are listed in Table~\ref{tab:e1_nf_defaults_all} and Table~\ref{tab:e1_nq_defaults_all} in the appendix.
\end{itemize}

For the TSFM zero-shot evaluation, we use only the price series because ChronosX’s covariate adapters are not pretrained. To keep the comparison fair, we therefore disable covariates for both Moirai and ChronosX, effectively reducing ChronosX to Chronos in this setting. For the fine-tuning comparison, calendar features are provided to all models. All models are trained and validated on the DE-LU splits defined in Section~\ref{sec:data} and are evaluated against the reference baseline using CRPS and the ES, along with additional metrics.

\subsection{E2: Hyperparameter Optimization}
\label{sec:hpo}

The second stage focuses on hyperparameter optimization for the two conventional deep learning methods: NF and NHITS+QRA. For each method we consider three architectural scales (tiny, small, base), resulting in six separate tuning rounds:

\begin{itemize}[leftmargin=1.5em]
    \item \textbf{NF}: three variants differing in Transformer width/depth and flow capacity.
    \item \textbf{NHITS+QRA}: three variants with different NHITS capacities and QRA design hyperparameters.
\end{itemize}

Hyperparameter search is performed on the DE-LU train/validation split using Optuna (NSGA-II), under the same realistic masking and price-only+calendar feature setup as in E1. We formulate tuning as a multi-objective optimization: for NF we minimize validation ES and ECE (see Appendix~\ref{sec:loss_fun_scores}), while for NHITS+QRA we minimize validation CRPS and ECE. The goal is to approximate a Pareto front, from which we select a balanced mid-tradeoff configuration. The hyperparameter search spaces are provided on the project’s GitHub page, and the selected tuned configurations are reported in Tables~\ref{tab:e2_nf_tuned_all} and~\ref{tab:e2_nq_tuned_all}.

\subsection{E3: Feature Selection}
\label{sec:features}
 
In the third stage we expand the input space beyond prices and calendar features to investigate the importance of additional covariates (market, supply, and cross-border variables). We select\footnote{Throughout this paper, the ``best'' configuration or variant denotes the one offering the most favorable trade-off between forecast accuracy, calibration, and computational cost (model size, training time, and required compute), rather than the one with the lowest error alone.}:
 
\begin{itemize}[leftmargin=1.5em]
    \item the best NF configuration from E2,
    \item the best NHITS+QRA configuration from E2,
    \item the best Moirai variant from E1,
    \item the best ChronosX variant from E1.
\end{itemize}

For each of the four reference models, we conduct a structured feature-selection procedure in which predefined feature groups and their combinations are evaluated. These groups are designed to reflect different sources of market information and are described below. For each time-varying covariate, we distinguish between two representations based on information availability at forecast time. The past-only representation corresponds to historically observed values available up to the forecast origin. In addition, we construct a corresponding future-only representation as a simple proxy for the forecast horizon, defined as the value of the same feature observed at the same hour exactly one week earlier. This design enables the models to exploit weekly seasonality while respecting realistic information constraints. Based on this construction, we define the following feature groups:

\begin{itemize}[leftmargin=1.5em]
    \item \textbf{R1}: CO\(_2\) allowances and load -- includes CO\(_2\) allowance prices and system load.
    \item \textbf{R2}: Gas and synthetic prices -- includes gas prices and synthetic electricity prices.
    \item \textbf{R3}: Generation mix -- includes non-renewable and renewable generation.
    \item \textbf{R4}: Cross-border trading and generation mix -- includes cross-border trading volumes together with non-renewable and renewable generation.
    \item \textbf{R5}: Load and generation mix -- includes system load together with non-renewable and renewable generation.
\end{itemize}

We perform feature selection programmatically via a structured feature-group evaluation procedure over predefined feature groups. At each step, we temporarily add one candidate group to the current feature set and assess the change in predictive accuracy using a Diebold-Mariano test with Newey-West heteroskedasticity- and autocorrelation-consis\-tent (HAC) correction. The loss differential is computed on CRPS for NHITS+QRA and on ES for all other models. The selected feature set for each model is the one that yields the best validation performance relative to the calendar-only baseline (CRPS for NHITS+QRA; ES otherwise).

For the Diebold--Mariano test, we use a Newey--West lag rule of $T^{0.25}$, a significance level of $\alpha = 0.05$, and a Student-$t$ reference distribution~\cite{dieboldComparingPredictiveAccuracy1995}. The Diebold--Mariano test is explained in more detail in the appendix in Section~\ref{sec:diebold_mariano}.

\subsection{E4: Learning Strategy Experiments (X-shot)}
\label{sec:xshot}

The fourth stage examines learning strategies under limited target-zone data (zero-shot, one-shot, few-shot). Because the set of features available across all zones is limited, we restrict E4 to the common feature set available for every zone. This group of experiments was conducted in parallel to E3 but uses a reduced input set consisting again of electricity prices and calendar features only. We reuse the same configurations as in E3, selected according to the criterion described in Section~\ref{sec:features}.

This set of experiments aims to quantify the amount of DE-LU--specific data required to adapt models trained on other bidding zones to the target zone, enabling a direct comparison between TSFMs and task-specific deep learning models across varying data regimes.

\section{Results and Discussion}
\label{sec:res}

When considering the task-specific models in Table~\ref{tab:e1_delu_crps_zero_shot_vs_finetuned}, no clear trend emerges. The results for the NF models are relatively close to each other, whereas NHITS+QRA shows greater variation. Interestingly, the largest model, denoted as base, performs worst. After hyperparameter tuning, however, this pattern is no longer observed; see Table~\ref{tab:e2_hp_tuning}. The results are then relatively similar across model variants. Overall, NHITS+QRA performs better than NF models.

Regarding the TSFMs, the zero-shot results are worse than the fine-tuned results, as expected. For Moirai, the results are very similar across all model sizes. For ChronosX, the differences are more pronounced. In the fine-tuned setting, the medium-sized model, denoted as base, performs worst compared with the smallest and largest variants. As the small variant performs within 0.23 CRPS of the largest, we proceed with ChronosX small for E3 and E4.

Finally, after hyperparameter tuning, NHITS+QRA achieves CRPS values comparable to those of the TSFMs in the zero-shot setting.

\begin{table}[pos=htbp]
  \centering
  \begin{threeparttable}
  \caption{Zero-shot vs.\ fine-tuned CRPS results for the DE-LU bidding zone. NF and NHITS+QRA were evaluated using fixed default settings.}
  \label{tab:e1_delu_crps_zero_shot_vs_finetuned}
  \small
  \renewcommand{\arraystretch}{1.15}
      \begin{tabular*}{\columnwidth}{@{\extracolsep{\fill}}llcc@{}}
      \multicolumn{4}{c}{\textbf{E1}} \\
      \toprule
      \multirow{2}{*}{Model} & \multirow{2}{*}{Size} & \multicolumn{2}{c}{Test CRPS} \\
      \cmidrule(lr){3-4}
      & & zero-shot & fine-tuned \\
      \midrule
      \multirow{3}{*}{NF}
      & tiny  & --    & 25.34 \\
      & small & --    & 24.80 \\
      & base  & --    & 24.65 \\
      \addlinespace[0.3em]
      \multirow{3}{*}{NQ}
      & tiny  & --    & 16.96 \\
      & small & --    & 27.99 \\
      & base  & --    & 41.90 \\
      \addlinespace[0.3em]
      \multirow{3}{*}{MO}
      & small & 17.70 & 14.91 \\
      & base  & 16.63 & 14.22 \\
      & large & 17.23 & 14.43 \\
      \addlinespace[0.3em]
      \multirow{3}{*}{CX}
      & small & 16.25 & 13.86 \\
      & base  & 15.84 & 15.60 \\
      & large & 15.76 & 13.63 \\
      \bottomrule
      \end{tabular*}
  \begin{tablenotes}[flushleft]
  \footnotesize
  \item NF (normalizing flow), NQ (NHITS+QRA), MO (Moirai), CX (ChronosX). Lower scores are better. `--' indicates no zero-shot value available.
  \end{tablenotes}
  \end{threeparttable}
\end{table}

As shown in Table~\ref{tab:e2_hp_tuning}, the NF model achieves its best performance at the medium model size (referred to as small), which we therefore adopt for all subsequent NF experiments.
For NHITS+QRA, the best performance is obtained with the largest configuration (denoted as base). However, given the substantial increase in model size relative to the more compact tiny variant and the comparatively small performance gap between the two, we proceed with the tiny configuration for E3 and E4.

\begin{table}[pos=htbp]
  \centering
  \begin{threeparttable}
  \caption{Results of the proposed models after hyperparameter-tuning on the DE--LU bidding zone.}
  \label{tab:e2_hp_tuning}
  \small
  \renewcommand{\arraystretch}{1.15}
      \begin{tabular*}{\columnwidth}{@{\extracolsep{\fill}}llc@{}}
      \multicolumn{3}{c}{\textbf{E2}} \\
      \toprule
      Model & Size & Test CRPS \\
      \midrule
      \multirow{3}{*}{NF}
      & tiny  & 25.41 \\
      & small & 24.23 \\
      & base  & 25.00 \\
      \addlinespace[0.3em]
      \multirow{3}{*}{NQ}
      & tiny  & 16.87 \\
      & small & 17.07 \\
      & base  & 16.57 \\
      \bottomrule
      \end{tabular*}
  \begin{tablenotes}[flushleft]
  \footnotesize
  \item NF (normalizing flow), NQ (NHITS+QRA). Lower scores are better.
  \end{tablenotes}
  \end{threeparttable}
\end{table}

As shown in Table~\ref{tab:e3_test_crps_groups_nhits_moirai}, NHITS+QRA benefits consistently from several additional feature groups. The inclusion of R3 yields a CRPS of 16.27, while R4 and R5 further reduce the score to 16.06 and 15.98, respectively. Among the evaluated feature groups, R5 therefore achieves the best performance and improves upon the tiny calendar-only NHITS+QRA reference in E1 by 0.89 CRPS.
In contrast, none of the remaining models improved. Moirai degrades sharply across every group it was evaluated on, from 14.22 to between 18.39 and 18.97, worse than its own zero-shot score in E1 (16.63). For NF--small and CX--small, no group produced a statistically significant validation improvement over the calendar-only baseline (Diebold--Mariano test), so no test-set evaluation was performed. These results suggest that performance is sensitive to the composition of the input feature set and that additional covariates do not uniformly translate into improved probabilistic accuracy.
A more exhaustive evaluation over all possible feature combinations would be required to fully validate these findings; such an analysis was not feasible within the available time and computational budget. We include only those feature groups whose inclusion leads to statistically significant differences in performance.

\begin{table}[pos=htbp]
\centering
\begin{threeparttable}
\caption{Test CRPS by feature group. Blank cells denote groups that were not evaluated on the test set because they did not yield a statistically significant improvement (or degraded performance) on the validation set.}
\label{tab:e3_test_crps_groups_nhits_moirai}
\small
\renewcommand{\arraystretch}{1.1}

\begin{tabular*}{\columnwidth}{@{\extracolsep{\fill}}lcccccc@{}}
\multicolumn{7}{c}{\textbf{E3}} \\
\toprule
Model & baseline & R1 & R2 & R3 & R4 & R5 \\
\midrule
NF--small & 24.23 & -- & -- & -- & -- & -- \\
NQ--tiny  & 16.87 & -- & -- & 16.27 & 16.06 & 15.98 \\
MO--base  & 14.22 & 18.39 & 18.97 & 18.63 & -- & -- \\
CX--small  & 13.86 & -- & -- & -- & -- & -- \\
\bottomrule
\end{tabular*}%

\begin{tablenotes}[flushleft]
  \footnotesize
  \item NF (normalizing flow), NQ (NHITS+QRA), MO (Moirai), CX (ChronosX). Results denote the test CRPS. Lower scores are better.
\end{tablenotes}
\end{threeparttable}
\end{table}

As depicted in Table~\ref{tab:cross_border_part_results}, the NF model shows a consistent improvement in probabilistic performance as additional data are introduced, with both CRPS and ES decreasing from zero-shot to one-shot and improving more substantially under few-shot adaptation. This pattern suggests that NF benefits directly from increased data availability and can effectively exploit cross-market information. 
NHITS+QRA and ChronosX, by contrast, are essentially flat across the three regimes: NHITS+QRA spans 13.54--13.60 and ChronosX 13.29--13.64, ranges we regard as indistinguishable given that all values come from single runs. The two models are therefore at parity in this setting; we do not read the 0.04 CRPS gap at few-shot in favor of NHITS+QRA as a meaningful ordering, and we note that ChronosX attains the single lowest value in the table (13.29, one-shot). For Moirai, performance deteriorates relative to its fully fine-tuned E1 configuration and remains between 16.49 and 16.84 irrespective of the number of DE-LU samples, indicating no benefit from cross-market pretraining in this setting.
The practically relevant observation is that cross-zone pretraining moves NHITS+QRA from 16.87 (E2) and 15.98 (E3) to 13.54--13.60, i.e.\ into the same range as ChronosX, while remaining a model of roughly $10^5$ parameters. Notably, despite ChronosX benefiting from a pretrained backbone, cross-zone pretraining enables the comparatively small NHITS+QRA model to reach a similar level of performance with only roughly \(10^5\) parameters, highlighting how competitive a well-calibrated task-specific model can become when trained on sufficient cross-market data.

\begin{table}[pos=htbp]
  \centering
  \begin{threeparttable}
  \caption{Results of the models trained or fine-tuned on the different European bidding zones with a varying number of samples from the DE-LU bidding zone.}
  \label{tab:cross_border_part_results}
  \small
  \renewcommand{\arraystretch}{1.15}
  \begin{tabular*}{\columnwidth}{@{\extracolsep{\fill}}llc@{}}
      \multicolumn{3}{c}{\textbf{E4}} \\
      \toprule
      Model & Strategy & Test CRPS \\
      \midrule
      \multirow{3}{*}{NF--small}
      & zero-shot & 22.92 \\
      & one-shot  & 22.89 \\
      & few-shot  & 19.97 \\
      \addlinespace[0.3em]
      \multirow{3}{*}{NQ--tiny}
      & zero-shot & 13.54 \\
      & one-shot  & 13.58 \\
      & few-shot  & 13.60 \\
      \addlinespace[0.3em]
      \multirow{3}{*}{MO--base}
      & zero-shot & 16.49 \\
      & one-shot  & 16.84 \\
      & few-shot  & 16.59 \\
      \addlinespace[0.3em]
      \multirow{3}{*}{CX--small}
      & zero-shot & 13.40 \\
      & one-shot  & 13.29 \\
      & few-shot  & 13.64 \\
      \bottomrule
      \end{tabular*}
  \begin{tablenotes}[flushleft]
  \footnotesize
  \item NF (normalizing flow), NQ (NHITS+QRA), MO (Moirai), CX (ChronosX). Zero-shot includes no samples from DE-LU. One-shot contains one window, while few-shot contains 23 windows. Lower scores are better.
  \end{tablenotes}
  \end{threeparttable}
\end{table}

Figure~\ref{fig:stacked_4} presents four fan charts for randomly selected forecast origins. These visualizations are intended for illustration only and are not suitable for assessing the calibration of individual models. For this purpose, PIT histograms or the ECE provide more appropriate evaluation tools.

\begin{figure*}[pos=h]
  \centering
  \begin{subfigure}{0.48\linewidth}
    \centering
    \includegraphics[width=\linewidth]{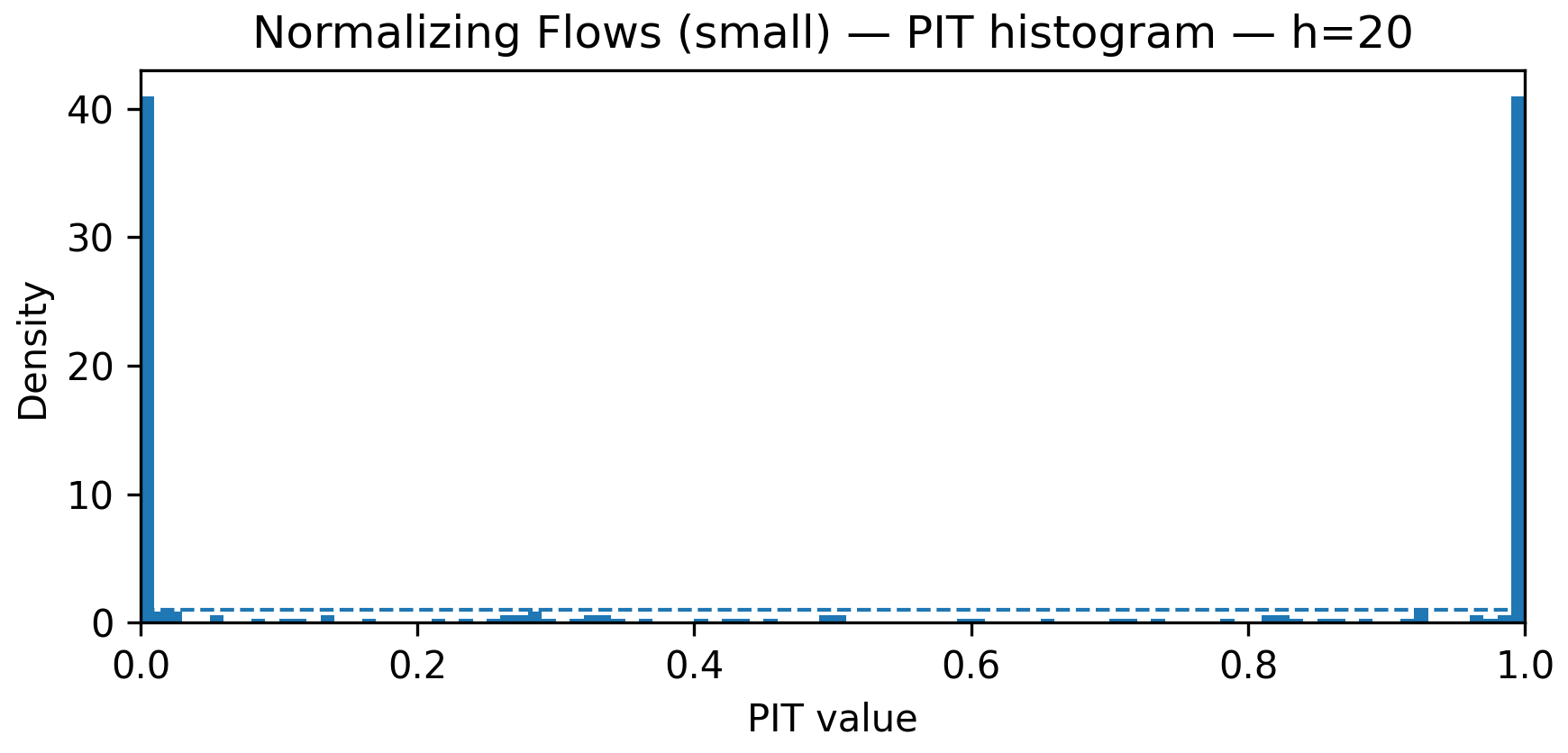}
    \caption{\textbf{E4}: NF-small few-shot}
    \label{fig:pit_nf_small}
  \end{subfigure}
  \hfill
  \begin{subfigure}{0.48\linewidth}
    \centering
    \includegraphics[width=\linewidth]{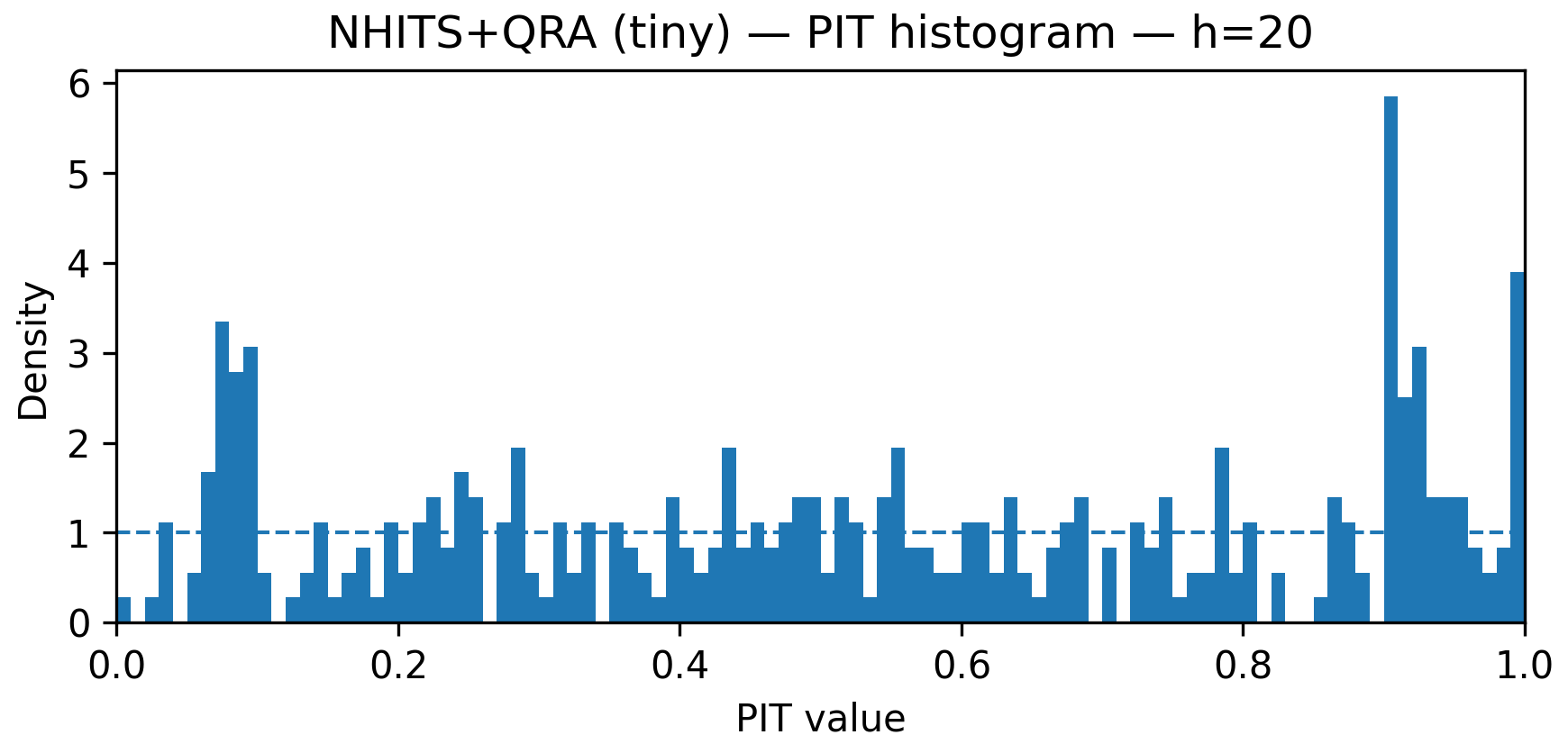}
    \caption{\textbf{E4}: NQ-tiny few-shot}
    \label{fig:pit_nq_tiny}
  \end{subfigure}
  \vspace{0.6em}
  \begin{subfigure}{0.48\linewidth}
    \centering
    \includegraphics[width=\linewidth]{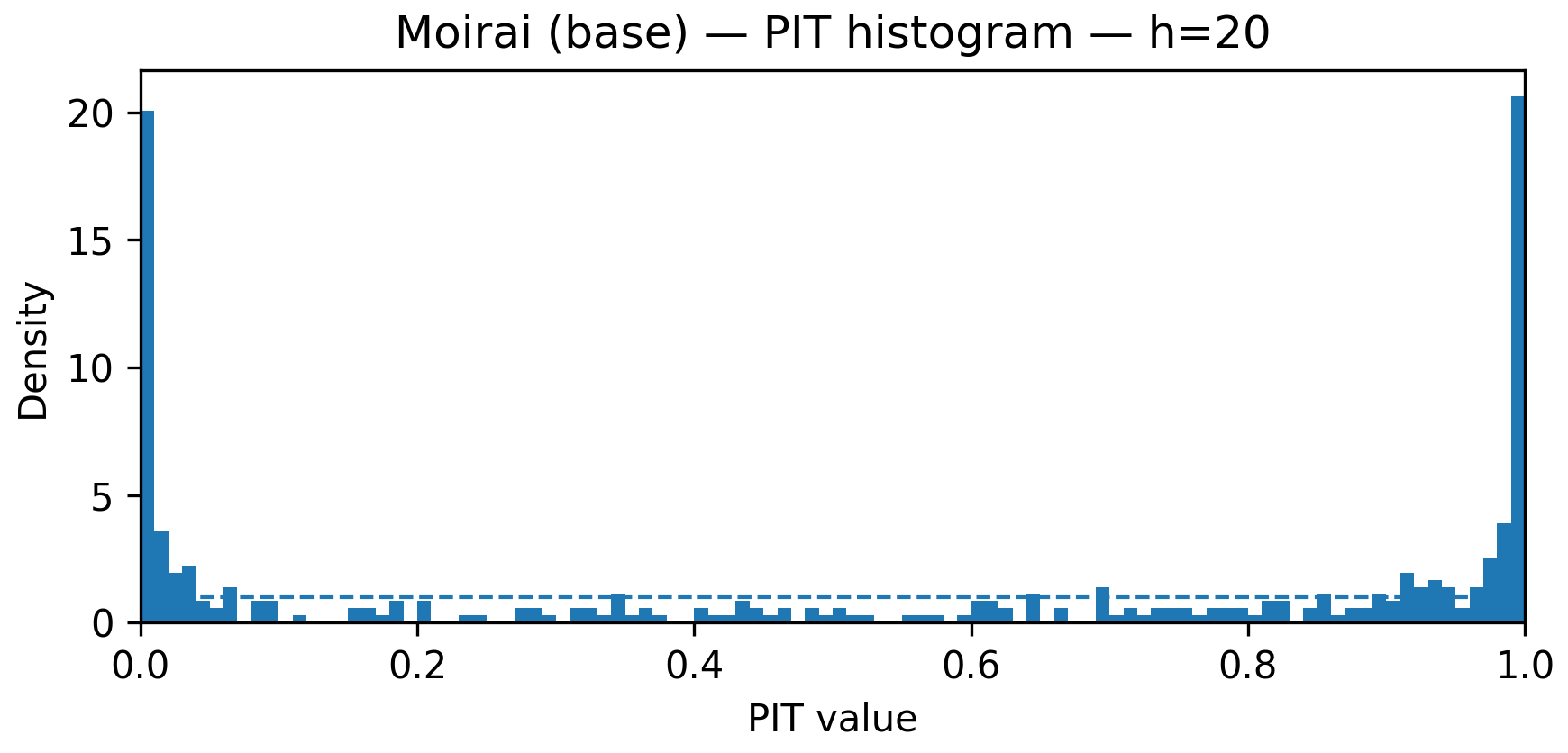}
    \caption{\textbf{E4}: MO-base few-shot}
    \label{fig:pit_mo_base}
  \end{subfigure}
  \hfill
  \begin{subfigure}{0.48\linewidth}
    \centering
    \includegraphics[width=\linewidth]{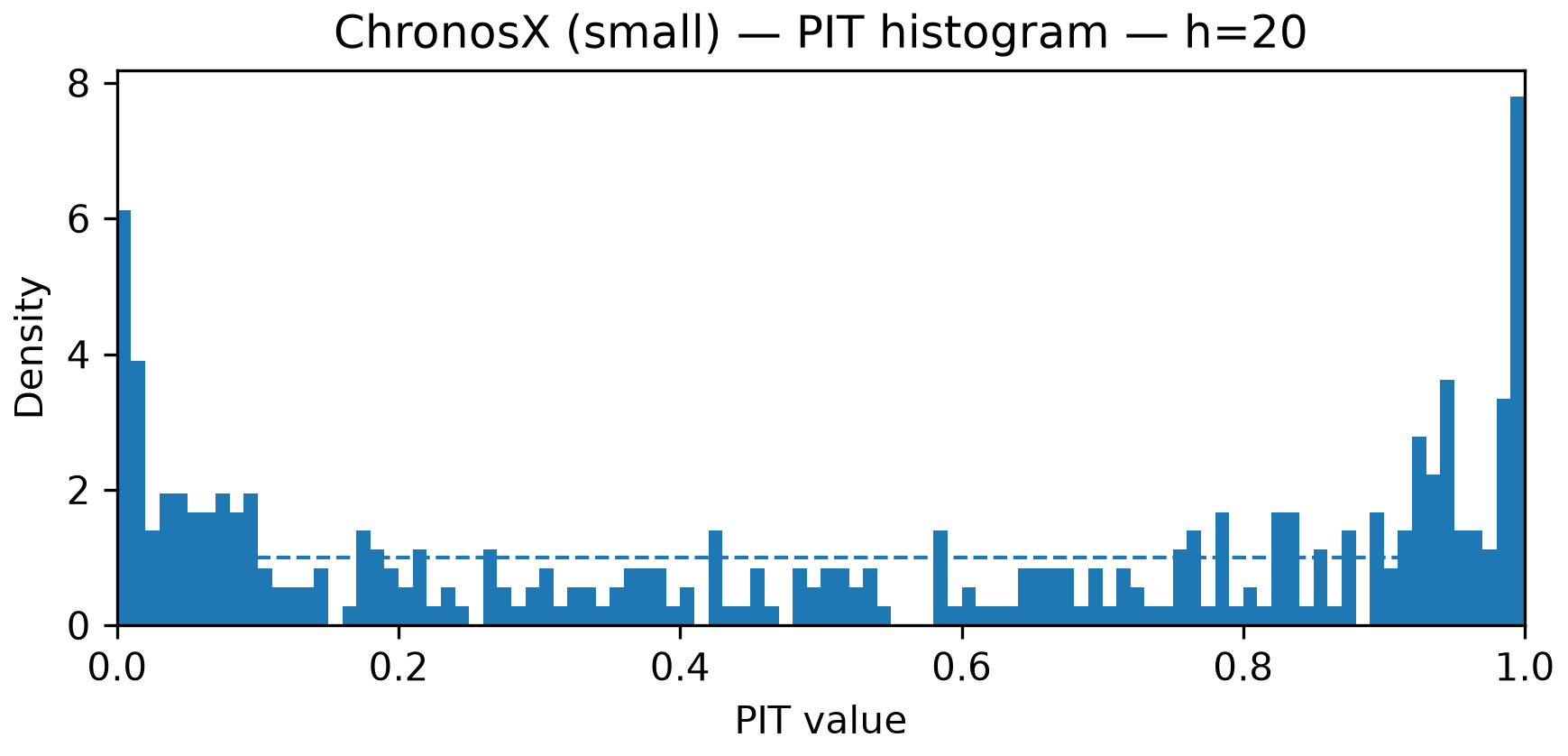}
    \caption{\textbf{E4}: CX-small few-shot}
    \label{fig:pit_cx_large}
  \end{subfigure}
\caption{PIT histograms for all models at forecast horizon 20. NF (normalizing flow), NQ (NHITS+QRA), MO (Moirai), CX (ChronosX).}
  \label{fig:pits_all_models}
\end{figure*}

Based on the ECE results reported in Tables~\ref{tab:e0_zero_shot_de_lu_mo_cx} and~\ref{tab:e1_full_finetune_delu_all_models} in the appendix, the TSFMs outperform the task-specific models. This pattern remains unchanged after tuning, as shown in Table~\ref{tab:e2_results_after_tuning} in the appendix. However, the results change in experiment E4; see Table~\ref{tab:e4_cross_border_full_results} and Figure~\ref{fig:pits_all_models}. Since the PIT histograms vary across forecast horizons, we show an illustrative example for $h=20$.

For the NF and Moirai models, the overconfident nature of the forecasts is clearly visible, with most of the probability mass concentrated near the boundaries of the plot. For ChronosX, we observe a bias, indicating that the model underestimates prices at the respective forecast horizon. NHITS+QRA shows the most evenly distributed probability mass and the lowest ECE in this setting. It is important to note, however, that this calibration advantage is partly architectural rather than a property of the underlying forecaster: the QRA head explicitly fits conditional quantiles on a held-out set via pinball loss with monotonicity enforcement, providing a form of post-hoc calibration by construction. The other three models, by contrast, rely on calibration emerging implicitly from their training objective (ES for NF, NLL for Moirai, and token-level cross-entropy for ChronosX). The comparison should therefore be read not as evidence that the NHITS backbone is intrinsically superior, but as evidence that explicit post-hoc calibration of a strong point forecaster is competitive with, and in this setting outperforms, pure likelihood-based or energy-score-based training of more expressive distributional models.

Overall, these results highlight that the models do not estimate uncertainty adequately, despite the fact that NF, NHITS+QRA, and Moirai were trained using strictly proper scoring rules: ES, pinball loss, and NLL, respectively~\cite{gneitingStrictlyProperScoring2007, chungPinballLossQuantile2021}. Strict propriety is a population-level property: in expectation, the true predictive distribution uniquely minimizes a strictly proper scoring rule. It does not, however, guarantee that a model fitted on finite data will recover this distribution or be empirically calibrated. In practice, finite training sets, model misspecification, imperfect optimization, and distribution shift between the training period (which includes the 2022 energy crisis) and the 2024 test period can all contribute to empirical miscalibration. Our findings are consistent with this and motivate the use of explicit calibration mechanisms on top of proper-scoring-rule training. The comparison further indicates that TSFMs are not uniformly better calibrated: in the data-richer cross-market setting (E4), the calibration ranking reverses in favor of NHITS+QRA, suggesting that the experimental regime plays a substantial role in determining which approach yields more reliable uncertainty estimates.

\subsection{Performance--Efficiency Trade-off}
\label{sec:efficiency}

We compare cost on three axes that are computable from the model configurations and our measured runs: model size, per-forecast compute, and training energy. Table~\ref{tab:perf_eff} summarizes all three alongside the corresponding accuracy in both adaptation regimes. Energy figures are training-phase measurements from Tables~\ref{tab:e1_energy_co2}, \ref{tab:energy_co2_feature_eval} and~\ref{tab:energy_co2_strategy_nf_nq_mo}; they were recorded on our hardware and should be read as order-of-magnitude indications. We report where each family wins, including where the task-specific models lose.

In the single-market setting (E1), the TSFMs are not the expensive option and we do not claim otherwise. Fine-tuning Moirai (base) consumed 0.37\,kWh against 0.89\,kWh for training NHITS+QRA (tiny) from scratch, and it achieved the better CRPS (14.22 against 16.96). ChronosX (small), the variant we carry forward into E3 and E4, cost 0.49\,kWh to fine-tune and reached 13.86, the second-best score in this stage. More striking still, the same model evaluated zero-shot, at 0.03\,kWh and no training at all, reached 16.25, which already improves on every configuration we trained from scratch. For a practitioner with one market, calendar features only, and no recurring retraining, the efficient choice is a pretrained model.

The ordering shifts under cross-zone pretraining. In E4, NHITS+QRA (tiny) consumed 1.06 to 1.21\,kWh against 10.37 to 10.97\,kWh for Moirai (base), a factor of roughly nine, while scoring 2.9 to 3.3 CRPS better and attaining the lower ECE (0.062 against 0.132). Against Moirai this is a dominance on both axes simultaneously, and it is the only such result in our study. ChronosX (small) is the harder comparison: it matches NHITS+QRA on accuracy (13.64 against 13.60, a gap we treat as parity) at a training cost of 6.19 to 7.39\,kWh. They are effectively tied in accuracy, while NHITS+QRA requires substantially less adaptation energy. The mechanism behind the gap to Moirai is straightforward: training cost scales with data volume, and a $10^{5}$ parameter model absorbs 15 zones of hourly data far more cheaply than a $10^{8}$ parameter Transformer. The same reasoning explains the feature-search costs in E3, where evaluating candidate groups cost 5.62\,kWh for NHITS+QRA (tiny), 1.69\,kWh for ChronosX (small), and 10.33\,kWh for Moirai (base).

Two structural differences persist across both regimes and do not depend on our measurements, as the upper block of Table~\ref{tab:perf_eff} shows. First, model size: the tuned NHITS+QRA (tiny) holds $1.15\times10^{5}$ parameters against $4.6\times10^{7}$ for ChronosX (small) and $9.1\times10^{7}$ for Moirai (base), factors of roughly 400 and 790, corresponding to deployment footprints of 0.46\,MB against 184\,MB and 364\,MB. Second, per-forecast compute: producing one 24-hour predictive distribution requires eight stochastic forward passes through the NHITS backbone (Table~\ref{tab:e2_nq_tuned_all}), on the order of $10^{6}$ floating-point operations, against roughly $10^{9}$ for a single pass through the parametric head of Moirai (base) and roughly $10^{11}$ for ChronosX (small), which autoregressively decodes 24 tokens for each of on the order of $10^{2}$ sampled trajectories. These gaps are invisible in wall-clock terms for a single forecast, and indeed our measured inference times slightly favor Moirai, but they set the marginal cost of scaling a deployment across zones and reforecasting intervals.

Finally, the adaptation energy we measure for the TSFMs excludes their pretraining. The Chronos authors report 63.1\,h on eight A100 GPUs for their largest model, on the order of 200\,kWh at typical device power, roughly an order of magnitude more than the entire E1 campaign reported here; Moirai does not report comparable figures. Amortized across all downstream users this is a small per-user cost and we do not charge it to the model. We note it because a comparison conducted on marginal adaptation energy alone attributes to architectural efficiency what is in part a transfer of cost outside the comparison.

We therefore do not claim that task-specific models are categorically more efficient. The defensible claim is narrower. NHITS+QRA dominates Moirai on both accuracy and energy in the cross-zone regime, matches ChronosX on accuracy there, remains two to three orders of magnitude smaller and several orders of magnitude cheaper per forecast throughout, and is the only model in our study that converts exogenous fundamentals into improved accuracy. The regimes in which it is the efficient choice are therefore precisely those identified in Sections~\ref{sec:features} and~\ref{sec:xshot}: repeated retraining, multiple markets, and informative exogenous features. Where none of these apply, a pretrained model used zero-shot or lightly fine-tuned is both cheaper and more accurate. The normalizing flow is efficient on neither axis and we do not advance it as a competitive option.

\begin{table}[pos=htbp]
  \centering
  \begin{threeparttable}
  \caption{Cost against accuracy in the two adaptation regimes. Energy is the training phase from Tables~\ref{tab:e1_energy_co2} and~\ref{tab:energy_co2_strategy_nf_nq_mo}; FLOPs are order-of-magnitude estimates per 24-hour predictive distribution.}
  \label{tab:perf_eff}
  \fontsize{7.5pt}{7.5pt}\selectfont
  \renewcommand{\arraystretch}{1.15}
    \begin{tabular*}{\columnwidth}{@{\extracolsep{\fill}}lrrrr@{}}
    \toprule
    & NF (small) & NQ (tiny) & MO (base) & CX (small) \\
    \midrule
    Parameters        & 3.5\,M & 115.3\,K & 91\,M & 46\,M \\
    Weights (fp32)    & 14\,MB & 0.46\,MB & 364\,MB & 184\,MB \\
    FLOPs / forecast  & $\sim\!10^{10}$ & $\sim\!10^{6}$ & $\sim\!10^{9}$ & $\sim\!10^{11}$ \\
    \midrule
    E1 energy (kWh)   & 1.07 & 0.89 & 0.37 & 0.49 \\
    E1 test CRPS      & 24.80 & 16.96 & 14.22 & 13.86 \\
    \addlinespace[0.25em]
    E4 energy (kWh)   & 10.10 & 1.21 & 10.41 & 6.92 \\
    E4 test CRPS      & 19.97 & 13.60 & 16.59 & 13.64 \\
    \bottomrule
    \end{tabular*}
  \begin{tablenotes}[flushleft]
  \footnotesize
  \item NF (normalizing flow), NQ (NHITS+QRA), MO (Moirai), CX (ChronosX). Sizes are those carried into E3 and E4. E1 rows use the default configurations of Table~\ref{tab:e1_delu_crps_zero_shot_vs_finetuned}; E4 rows the few-shot setting.
  \end{tablenotes}
  \end{threeparttable}
\end{table}

\subsection{Limitations}
\label{sec:limitations}

Several limitations bound the strength of our conclusions.

\textbf{Single-seed runs.} Every number reported in this paper comes from one training run per configuration. We therefore cannot separate systematic differences from initialization variance, and we deliberately refrain from declaring winners where models differ by less than roughly 0.5 CRPS. This affects the NHITS+QRA/ChronosX comparison in E4 most directly, where the models are separated by 0.04-0.31 CRPS across regimes. The within-model spread of NHITS+QRA across the three E4 regimes (0.06 CRPS), which we attribute to noise, provides an informal lower bound on the resolution of our comparisons.

\textbf{No significance testing on test-set comparisons.} Die\-bold--Mariano testing is applied during feature selection on the validation set, but the headline model comparisons in Tables~\ref{tab:e1_delu_crps_zero_shot_vs_finetuned}--\ref{tab:cross_border_part_results} are unaccompanied point estimates. Establishing the reported orderings would require either multiple seeds or DM tests on test-set loss differentials.

\textbf{Single evaluation zone.} Although E4 uses 14 additional bidding zones, they serve exclusively as pretraining data; all evaluation is on DE-LU in 2024. Whether the feature-responsiveness gap between NHITS+QRA and the TSFMs generalizes to less interconnected or less renewable-heavy zones is untested.

\textbf{Baselines.} We compare against four transparent statistical baselines but not against LEAR or DNN-EPF~\cite{lagoForecastingDayaheadElectricity2021a}, two existing reference models in this field. We note that our strongest naive baseline, Same-Hour (Last 28 Days) at 19.90 CRPS (Table~\ref{tab:baseline_crps}), outperforms every normalizing flow configuration we trained; the NF results should accordingly be read as a negative result for that architecture under our training budget rather than as a competitive reference point.

\textbf{Unexplained configuration effects.} Two results in our tables warrant caution. The default (untuned) NHITS+QRA configurations degrade monotonically with size in E1 (16.96/\allowbreak{}27.99/\allowbreak{}41.90 for tiny/small/base), a pattern that disappears entirely after tuning; we attribute this to an interaction between the enlarged stochastic ensemble and the LASSO quantile fits at the default learning rate rather than to model capacity, but we have not isolated the cause. Relatedly, NHITS+QRA (tiny) pretrained on other zones with no DE-LU data (13.54, E4) outperforms the same architecture tuned directly on DE-LU (16.87, E2); the most plausible explanation is the far larger training volume available across 15 zones, but we cannot rule out differences in the QRA calibration split between the two pipelines.

\section{Conclusion}
\label{sec:sum}

In this work, we compare two TSFMs, Moirai and ChronosX, against two task-specific probabilistic forecasters, NHITS+QRA and NF, for day-ahead electricity prices on the DE-LU market. Our results indicate that TSFMs achieve strong performance in terms of CRPS, ES, and probabilistic calibration assessed via PIT histograms when fine-tuned on prices and calendar features alone, outperforming task-specific deep learning models. This ordering does not survive unchanged into the other regimes we study. NHITS+QRA is the only model that improves when supplied with exogenous fundamentals, while Moirai degrades below its own zero-shot performance; and under cross-market pretraining with limited target-zone data, NHITS+QRA reaches parity with ChronosX. It does so with roughly three orders of magnitude fewer parameters and, in that regime, about one ninth of the training energy of Moirai.

We are deliberately cautious about the strength of these statements. All results derive from single runs on a single evaluation zone, and several of the differences we discuss are small enough that we treat them as parity rather than as orderings; Section~\ref{sec:limitations} sets out these constraints in full. The normalizing flow, in particular, fails to beat a same-hour empirical baseline under our training budget and should not be read as a competitive reference.

These findings highlight the need to carefully weigh computational cost against marginal performance gains. Non-TSFMs may also offer practical advantages: easier distributed computing and better data privacy. Because the entire training-to-test pipeline can be controlled, they additionally provide stronger assurances against data leakage and a more realistic assessment of model performance.

Looking forward, our work motivates further research in two directions. First, the design of loss functions and architectures that target well-calibrated probabilistic forecasts more directly. Second, the combined use of cross-zone data and exogenous features to improve forecast quality across markets, particularly in renewable-dominated electricity systems. Taken together with the observation in Section~\ref{sec:res} that the post-hoc calibration provided by QRA performs competitively when more data is added, this motivates a research direction in which expressive distributional models are explicitly paired with dedicated calibration heads, rather than relying on proper-scoring-rule training alone. In addition, future work could explore the use of newer TSFMs, such as Chronos-2~\cite{ansariChronos2UnivariateUniversal2025}, Moirai 2.0~\cite{liuMoirai20When2025}, or Moirai-MoE~\cite{pmlr-v267-liu25an}.


\section*{Acknowledgements}
We would like to acknowledge the insightful and stimulating discussions with Prof.\ Dr.-Ing.\ Giovanni De~Carne and Muhammad Abdullah Malik.

We also gratefully acknowledge funding from the \allowbreak{} Helmholtz Association and the Networking Fund through Helmholtz AI under Grant No. VH-NG-1727.

This work was performed on three computing resources: 
\begin{itemize}
    \item The authors acknowledge support by the state of Baden-Württemberg through bwHPC.
    \item This work is supported by the Helmholtz Association Initiative and Networking Fund on the HAI\-CORE@KIT partition.
    \item The authors gratefully acknowledge the computing time provided on the high-performance computer HoreKa by the National High-Performance Computing Center at KIT (NHR@KIT). This center is jointly supported by the Federal Ministry of Education and Research and the Ministry of Science, Research and the Arts of Baden-Württemberg, as part of the National High-Performance Computing (NHR) joint funding program (https://www.nhr-verein.de/en/our-partners). HoreKa is partly funded by the German Research Foundation (DFG).
\end{itemize}







\appendix

\section*{Data Licensing and Terms of Use}
\label{sec:license}

The data provided by the Energy-Charts API is licensed under the CC BY 4.0 license. Proper attribution to Energy-Charts.info as the source is required.
The Fraunhofer-Gesell\-schaft retains full copyright over the content provided on the Energy Charts website. Downloading or printing of publications is permitted for personal use and for the purpose of reporting on the Fraunhofer-Gesellschaft and its institutes, in accordance with specified usage conditions. Any further use, especially commercial utilization and distribution, generally requires written permission. Requests should be directed to 
Fraunhofer-Institut fuer Solare Energiesysteme ISE
Heidenhofstr. 2
79110 Freiburg.
For full details on the terms of use and licensing, please refer to the official \href{https://www.energy-charts.info/publishing-notes.html?l=en&c=DE}{publishing notes} and \href{https://api.energy-charts.info/}{API licensing}.

\section*{Data Availability}
\label{sec:data_availability}
The electricity-market data used in this study are publicly available through the Energy-Charts API. The source code, model configurations, and experiment scripts required to reproduce the reported experiments will be made publicly available upon publication.

\section{Supplementary Figures}
\label{sec:supp_figures}
\begin{figure*}[t]
  \centering

  \begin{subfigure}{0.48\textwidth}
    \centering
    \includegraphics[width=\linewidth]{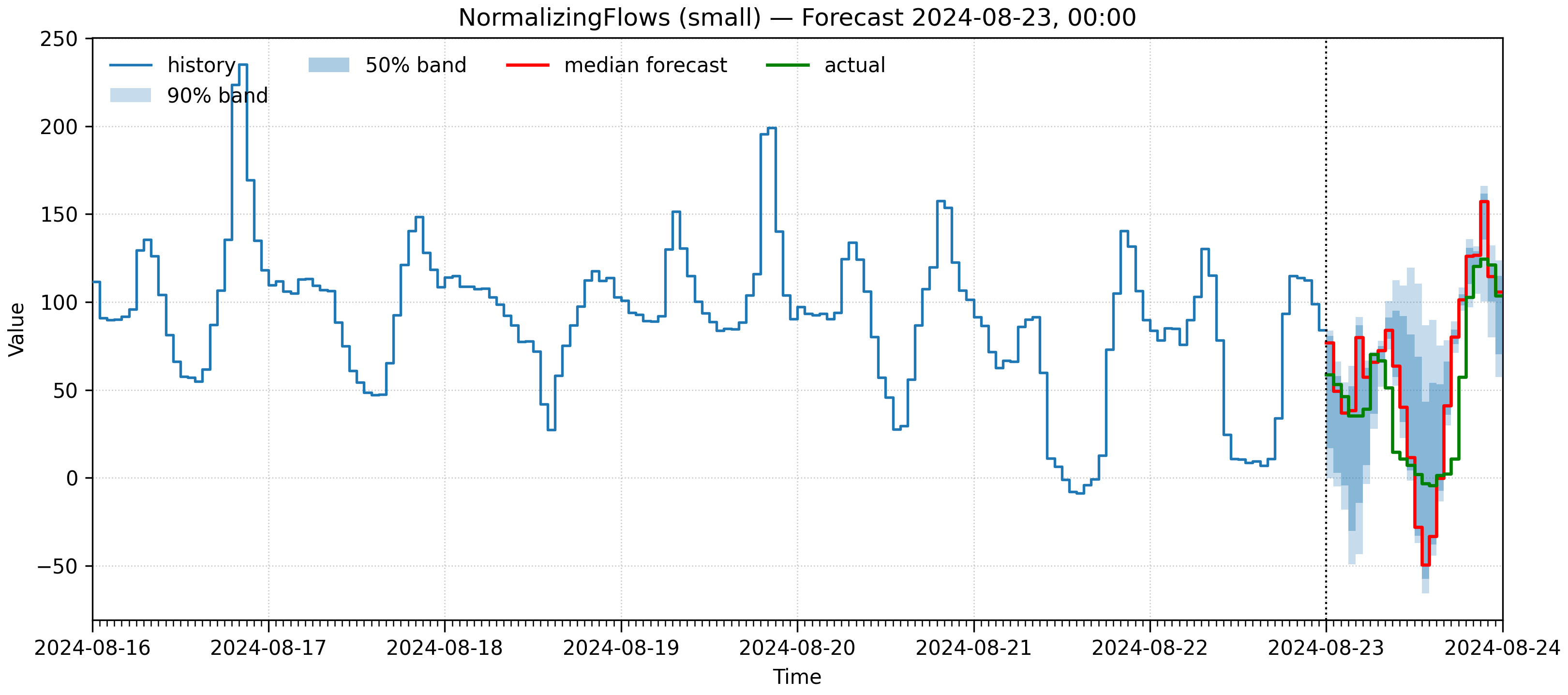}
    \caption{\textbf{E4}: NF-small few-shot}
    \label{fig:stack_1}
  \end{subfigure}
  \hfill
  \begin{subfigure}{0.48\textwidth}
    \centering
    \includegraphics[width=\linewidth]{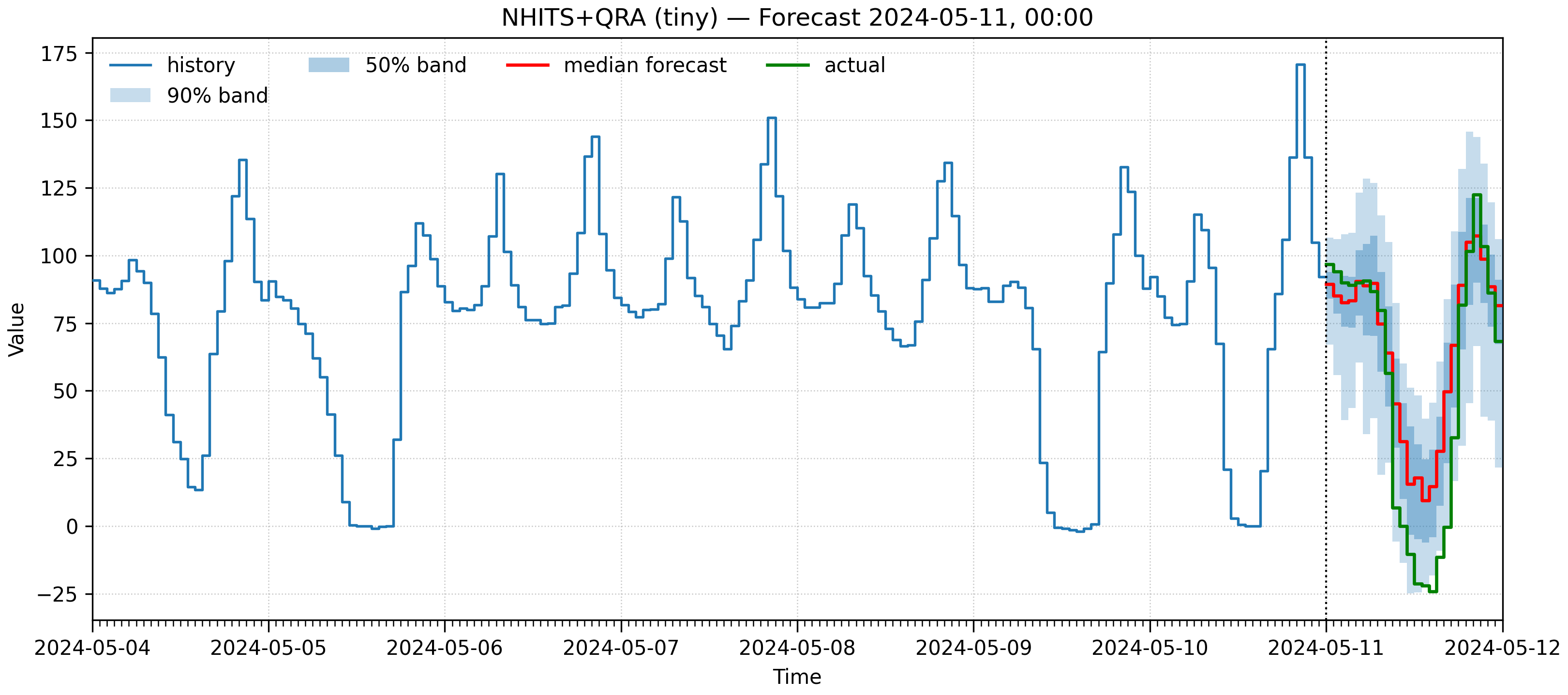}
    \caption{\textbf{E4}: NQ-tiny few-shot}
    \label{fig:stack_2}
  \end{subfigure}

  \vspace{0.8em}

  \begin{subfigure}{0.48\textwidth}
    \centering
    \includegraphics[width=\linewidth]{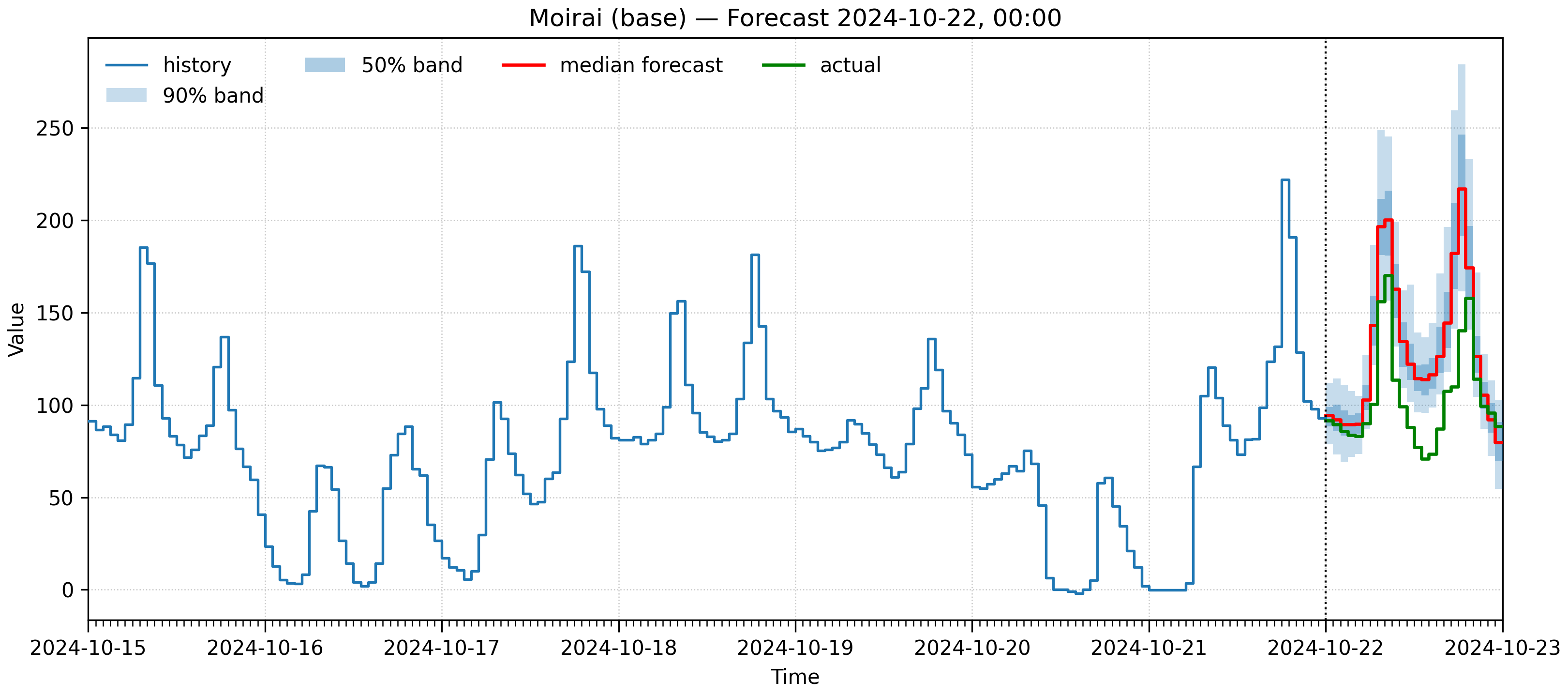}
    \caption{\textbf{E4}: MO-base few-shot}
    \label{fig:stack_3}
  \end{subfigure}
  \hfill
  \begin{subfigure}{0.48\textwidth}
    \centering
    \includegraphics[width=\linewidth]{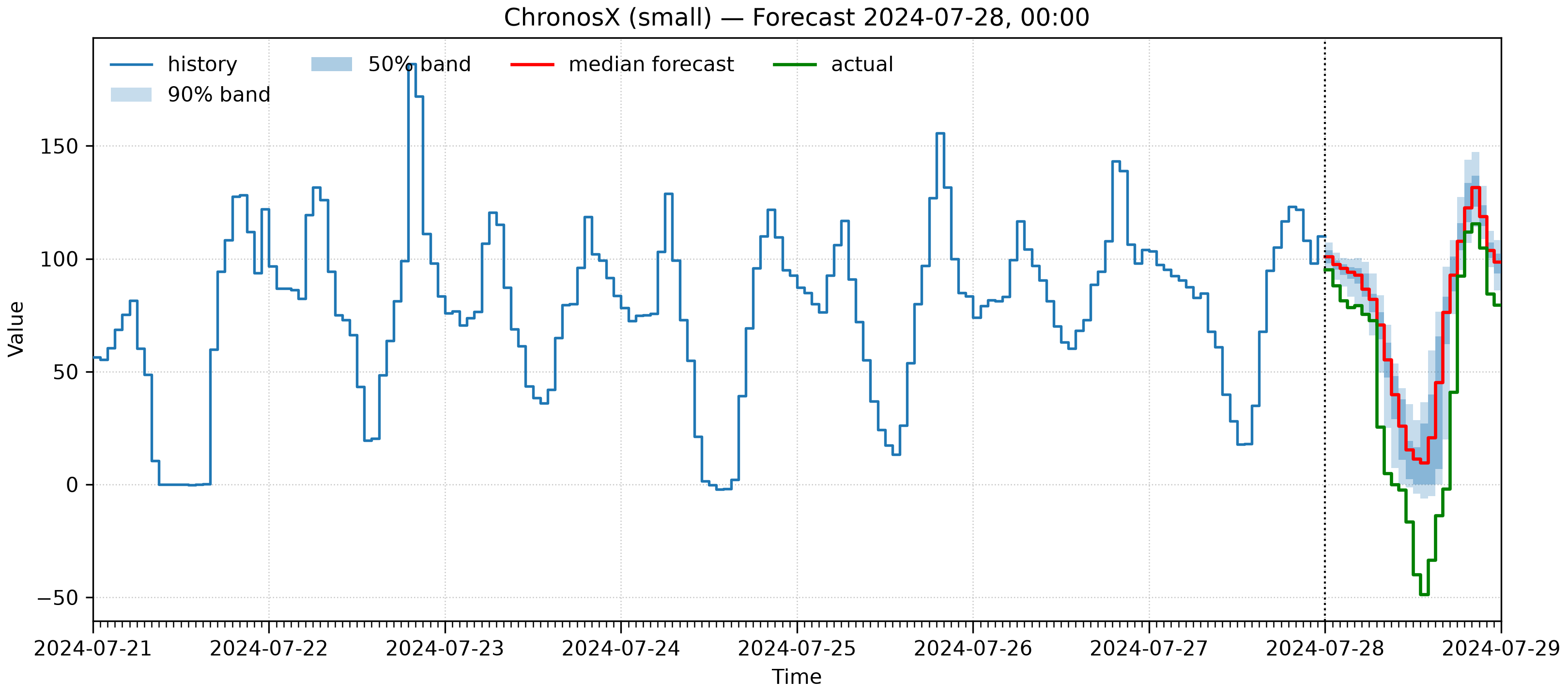}
    \caption{\textbf{E4}: CX-small few-shot}
    \label{fig:stack_4}
  \end{subfigure}

\caption{Fan charts for all models at different origins. NF (normalizing flow), NQ (NHITS+QRA), MO (Moirai), CX (ChronosX).}
  \label{fig:stacked_4}
\end{figure*}

Figure~\ref{fig:stacked_4} shows illustrative fan charts for all four model families at randomly selected forecast origins.

\section{Loss Functions and Scores}
\label{sec:loss_fun_scores}

\subsection{CRPS}

The CRPS measures the quality of a probabilistic forecast with cumulative distribution function (CDF) $F$ relative to an observed realization $y$. It is a strictly proper scoring rule~\cite{gneitingStrictlyProperScoring2007}, meaning that its expected value is minimized if and only if the forecast distribution $F$ coincides with the true underlying distribution. The CRPS is defined as
\[
\mathrm{CRPS}(F,y)
=
\int_{-\infty}^{\infty}\bigl(F(u)-\mathds{1}\{u\ge y\}\bigr)^2\,\mathrm{d}u.
\]

An equivalent representation is
\[
\mathrm{CRPS}(F,y)
=
\mathbb{E}|Y-y|
-\frac{1}{2}\,\mathbb{E}|Y-Y'|,
\]
where $Y$ and $Y'$ are independent random variables with distribution function $F$ and finite first moment~\cite{gneitingStrictlyProperScoring2007}.

\subsection{Energy Score (ES)}
The ES is the multivariate generalization of the CRPS. For a predictive distribution $F$ on $\mathbb{R}^m$ and an observed realization $\mathbf{y}\in\mathbb{R}^m$, it is defined as
\[
\mathrm{ES}(F,\mathbf{y})
=
\mathbb{E}_{F}\|\mathbf{Y}-\mathbf{y}\|^{\beta}
-\frac{1}{2}\,\mathbb{E}_{F}\|\mathbf{Y}-\mathbf{Y}'\|^{\beta},
\]
where $\mathbf{Y},\mathbf{Y}' \stackrel{\text{i.i.d.}}{\sim} F$~\cite{gneitingStrictlyProperScoring2007}.

\subsection{Expected Calibration Error}

The ECE measures the deviation from average calibration. It is defined as
\[
\mathrm{ECE}
=
\frac{1}{m}\sum_{j=1}^m \left|\hat p^{\mathrm{obs}}(p_j) - p_j\right|,
\]
where $m$ denotes the number of quantile levels at which the model is evaluated, $p_j$ is the target quantile level, and $\hat p^{\mathrm{obs}}(p_j)$ is the observed average calibration at level $p_j$~\cite{chungPinballLossQuantile2021}.

\subsection{Diebold--Mariano and Newey--West}
\label{sec:diebold_mariano}
The Diebold--Mariano (DM) test~\cite{dieboldComparingPredictiveAccuracy1995} compares two forecasting methods by testing whether they have equal expected predictive loss. For a loss or proper scoring rule \(L(\cdot)\) (e.g., CRPS or ES), define the loss differential
\[
d_t = L_t^{(A)} - L_t^{(B)}, \qquad t=1,\dots,T,
\]
and test \(H_0:\ \mathbb{E}[d_t]=0\). Under covariance stationarity of \(\{d_t\}\), the sample mean \(\bar d\) is asymptotically normal. Since \(d_t\) may be serially correlated, inference uses a HAC estimate of the long-run variance, commonly Newey--West~\cite{neweySimplePositiveSemiDefinite1987} with Bartlett weights:
\begin{gather*}
\widehat{\sigma}^2_{\mathrm{LR}}
= \hat\gamma_0 + 2\sum_{j=1}^{m} w_j\,\hat\gamma_j,
\qquad
w_j = 1-\frac{j}{m+1},
\\
\widehat{\mathrm{Var}}(\bar d)
= \frac{\widehat{\sigma}^2_{\mathrm{LR}}}{T}.
\end{gather*}
The DM statistic is
\[
\mathrm{DM}=\frac{\bar d}{\sqrt{\widehat{\mathrm{Var}}(\bar d)}} \;\xrightarrow{d}\; \mathcal{N}(0,1).
\]
In finite samples, it is common to use the small-sample adjustment of \citet{harveyTestingEqualityPrediction1997}
and compare against Student's \(t\) critical values with \(T-1\) degrees of freedom.

\section{Synthetic Price}
\label{sec:SyntheticPrice}
The synthetic price is a feature-engineered variable that combines gas prices and CO\textsubscript{2} emission allowances (Equation~\ref{eq:synthetic_price_formula}) to approximate electricity price formation. The formula is adapted from an explanation provided in a Fraunhofer Energy Talks video ~\cite{fraunhoferise43EnergyChartsTalks2025}. This feature provides an estimate of average spot prices in the Day-Ahead Auction (volume-weighted for the DE-LU zone), offering insights into market dynamics and price-setting mechanisms. Electricity prices are strongly influenced by gas prices and CO\textsubscript{2} emission allowances, reflecting their role in market price formation. The dependence on gas prices arises from the merit-order principle, as the market price is often set by the most expensive power plant required to meet demand, which may be a gas-fired plant. The efficiency of modern combined-cycle gas turbine (CCGT) power plants is typically above 55\%. Deviations from this simplified relationship may arise from profit margins, additional costs, and lower efficiencies of older power plants. We therefore assume an efficiency of 55\% in the synthetic-price calculation.

\begin{equation}
SP = \frac{GP}{\eta} + EI \cdot \frac{CP}{1000}
\label{eq:synthetic_price_formula}
\end{equation}

where $SP$ is the synthetic price (EUR/MWh), $GP$ the gas price (EUR/MWh), $\eta$ the plant efficiency, $EI$ the emission intensity, and $CP$ the CO$_2$ certificate price (EUR/tCO$_2$). We set $\eta = 0.55$ and $EI = 400\,\mathrm{gCO_2/kWh}$.

\FloatBarrier
\section{Baseline}
\label{sec:baselines}

To compare the performance of the proposed probabilistic models, we implement four
simple, fully transparent baselines. All baselines operate independently
on each time series, use only information
available strictly before the forecast origin, and are evaluated with the CRPS via ensemble samples. A common base class reads the training and test
CSV files, aligns timestamps, and provides utilities for index resolution,
historical lookup, and CRPS aggregation.

\paragraph{Same-Hour Empirical Distribution (Last 28 Days)}
For a forecast time $t$ with hour of day $h(t)$, we collect the observations
obtained at the same hour on the preceding 28 distinct calendar days:
\[
\mathcal{E}_{t}
= \bigl\{\, y_{t-d}\;|\;d\in\{1,\dots,28\},\;
\operatorname{hour}(t-d)=h(t) \,\bigr\}.
\]
The 28 values constitute an empirical ensemble on which the CRPS is
computed. The lookup proceeds backwards in calendar time until 28 distinct
days are found; duplicate records caused by daylight-saving-time (DST) shifts
are removed (keeping the first occurrence). If fewer than 28 past values exist
(e.g., at the very beginning of a series) the forecast is omitted and logged.

\paragraph{Same-Hour Empirical Distribution\\(7 Days + 12 Months)}
This baseline augments the short-term window with an annual seasonal component.
For each $t$ we build
\[
\begin{aligned}
\mathcal{E}_{t} &=
\underbrace{\bigl\{\, y_{t-d}\;|\;d\in\{1,\dots,7\},
\operatorname{hour}(t-d)=h(t) \,\bigr\}}_{\text{last 7 days}} \\
&\qquad\cup\;
\underbrace{\bigl\{\, y_{t-\text{1 month}},\dots,
y_{t-\text{12 months}}\;|\;
\operatorname{hour}(\cdot)=h(t) \,\bigr\}}_{\text{same day of prior 12 months}} .
\end{aligned}
\]
The two subsets are gathered using calendar-based offsets; missing dates are
skipped until the required count is reached, and DST duplicates are handled as
described above. The resulting $7+12$ samples form the empirical ensemble for
the CRPS.

\paragraph{Bootstrap of Electricity Prices}
A point forecast is obtained by lagging the target series by a fixed calendar
offset $L$ (typically one week):
\[
\hat{y}_{t}=y_{t-L},
\]
with a $\pm1$-hour fallback to cope with DST boundaries. During the training
period we compute one-step-ahead residuals
$\varepsilon_{\tau}=y_{\tau}-\hat{y}_{\tau}$ and store the empirical distribution
of these residuals. At test time we draw $M$ i.i.d. residuals
$\varepsilon^{(m)}$ and form bootstrap samples
\[
\tilde{y}^{(m)}_{t}= \hat{y}_{t} + \varepsilon^{(m)},\qquad m=1,\dots,M,
\]
which define the ensemble used for the CRPS. Residuals are drawn only
from timestamps strictly preceding the train-test split to avoid leakage; DST
duplicates are averaged.

\paragraph{Bootstrap of Synthetic Prices}
The same bootstrap scheme is applied, but the point forecast is taken from a
reference signal $x$ (the engineered \texttt{synthetic\_price}) rather than from
the target series:
\[
\hat{y}_{t}=x_{t-L}.
\]
Residuals are computed against the true target in the training period, then
resampled and added to $\hat{y}_{t}$ exactly as in the previous baseline. This approach tests how well a simple transformation of market drivers can capture the
target distribution when combined with the empirical error structure.

Among the four baselines, the \emph{Same-Hour (Last 28 Days)} model achieves the
lowest CRPS on the DE-LU test set and is therefore adopted as the
reference baseline in all subsequent experiments as seen in Table~\ref{tab:baseline_crps}.

\begin{table}[t]
\centering
\caption{Baseline test CRPS on DE-LU.}
\label{tab:baseline_crps}
\setlength{\tabcolsep}{6pt}
\renewcommand{\arraystretch}{1.15}
\begin{tabular}{@{} l r @{}}
\toprule
\textbf{Baseline} & \textbf{Test CRPS} \\
\midrule
28 d (hour)          & 19.90 \\
7 d + 12 m (hour)    & 21.10 \\
Boot (real)          & 30.16 \\
Boot (synth.)        & 30.16 \\
\bottomrule
\end{tabular}
\end{table}

\FloatBarrier
\clearpage
\section{Supplementary Results}
\begin{table}[pos=h]
  \centering
  \begin{threeparttable}
  \caption{Zero-shot results for the DE-LU bidding zone.}
    \label{tab:e0_zero_shot_de_lu_mo_cx}
  \small
  \setlength{\tabcolsep}{10pt}
  \renewcommand{\arraystretch}{1.15}
      \begin{tabular}{l l c c c }
      \multicolumn{5}{c}{\textbf{E1}} \\
      \toprule
      Model & Size & Test ECE & Test CRPS & Test ES \\
      \midrule
      \multirow{3}{*}{MO}
      & small & 0.076 & 17.70 & 19.02 \\
      & base  & 0.047 & 16.63 & 18.13 \\
      & large & 0.062 & 17.23 & 19.01 \\
      \addlinespace[0.3em]
      \multirow{3}{*}{CX}
          & small & 0.063 & 16.25 & 17.51 \\
          & base  & 0.057 & 15.84 & 17.02 \\
          & large & 0.060 & 15.76 & 16.94 \\
      \bottomrule
      \end{tabular}
  \begin{tablenotes}[flushleft]
  \footnotesize
  \item MO (Moirai), CX (ChronosX). Lower scores are better.
  \end{tablenotes}
  \end{threeparttable}
\end{table}

\begin{table}[pos=h]
  \centering
  \begin{threeparttable}
  \caption{Full fine-tuning results for the DE-LU bidding zone.}
  \label{tab:e1_full_finetune_delu_all_models}
  \small
  \setlength{\tabcolsep}{10pt}
  \renewcommand{\arraystretch}{1.15}

      \begin{tabular}{l l c c c }
      \multicolumn{5}{c}{\textbf{E1}} \\
      \toprule
      Model & Size & Test ECE & Test CRPS & Test ES \\
      \midrule

      \multirow{3}{*}{NF}
      & tiny  & 0.145 & 25.34 & 24.88 \\
      & small & 0.138 & 24.80 & 24.70 \\
      & base  & 0.132 & 24.65 & 24.23 \\
      \addlinespace[0.3em]

      \multirow{3}{*}{NQ}
      & tiny  & 0.112 & 16.96 & 16.82 \\
      & small & 0.143 & 27.99 & 29.26 \\
      & base  & 0.134 & 41.90 & 31.74 \\
      \addlinespace[0.3em]

      \multirow{3}{*}{MO}
      & small & 0.057 & 14.91 & 16.11 \\
      & base  & 0.042 & 14.22 & 15.55 \\
      & large & 0.046 & 14.43 & 15.80 \\
      \addlinespace[0.3em]

      \multirow{3}{*}{CX}
      & small & 0.076 & 13.86 & 14.93 \\
      & base  & 0.066 & 15.60 & 16.60 \\
      & large & 0.057 & 13.63 & 14.48 \\
      \bottomrule
      \end{tabular}
  \begin{tablenotes}[flushleft]
  \footnotesize
  \item Full fine-tuning results on the DE-LU bidding zone using default (non-tuned) configurations. NF (normalizing flow), NQ (NHITS+QRA), MO (Moirai), CX (ChronosX). Lower scores are better.
  \end{tablenotes}
  \end{threeparttable}
\end{table}

\begin{table}[pos=h]
  \centering
  \begin{threeparttable}
  \caption{Results of the proposed models after hyperparameter-tuning on the DE-LU bidding zone.}
  \label{tab:e2_results_after_tuning}
  \small
  \setlength{\tabcolsep}{10pt}
  \renewcommand{\arraystretch}{1.15}
      \begin{tabular}{l l c c c }
      \multicolumn{5}{c}{\textbf{E2}} \\
      \toprule
      Model & Size & Test ECE & Test CRPS & Test ES \\
      \midrule
      \multirow{3}{*}{NF}
      & tiny & 0.150 & 25.41 & 25.13 \\
      & small & 0.125 & 24.23 & 24.00 \\
      & base & 0.134 & 25.00 & 25.31 \\
      \addlinespace[0.3em]
      \multirow{3}{*}{NQ}
      & tiny & 0.107 & 16.87 & -- \\
      & small  & 0.111 & 17.07 & -- \\
      & base & 0.107 & 16.57 & -- \\
      \bottomrule
      \end{tabular}
  \begin{tablenotes}[flushleft]
  \footnotesize
  \item NF (normalizing flow), NQ (NHITS+QRA), MO (Moirai), CX (ChronosX). Lower scores are better.
  \end{tablenotes}
  \end{threeparttable}
\end{table}

\begin{table}[pos=h]
  \centering
  \begin{threeparttable}
  \caption{Results of the models trained or fine-tuned on the different European bidding zones with a varying number of samples from the DE-LU bidding zone.}
  \label{tab:e4_cross_border_full_results}
  \small
  \setlength{\tabcolsep}{7pt}
  \renewcommand{\arraystretch}{1.15}
      \begin{tabular}{l l c c c }
      \multicolumn{5}{c}{\textbf{E4}} \\
      \toprule
      Model & Strategy & Test ECE & Test CRPS & Test ES \\
      \midrule
      \multirow{3}{*}{NF-small}
      & zero-shot & 0.137 & 22.92 & 22.69 \\
      & one-shot & 0.136 & 22.89 & 22.67 \\
      & few-shot & 0.083 & 19.97 & 19.48 \\
      \addlinespace[0.3em]
      \multirow{3}{*}{NQ-tiny}
      & zero-shot & 0.061 & 13.54 & -- \\
      & one-shot  & 0.062 & 13.58 & -- \\
      & few-shot & 0.062 & 13.60 & -- \\
      \addlinespace[0.3em]
      \multirow{3}{*}{MO-base}
      & zero-shot & 0.136 & 16.49 & 17.21 \\
      & one-shot  & 0.132 & 16.84 & 17.55 \\
      & few-shot & 0.132 & 16.59 & 17.39 \\
      \addlinespace[0.3em]
      \multirow{3}{*}{CX-small}
      & zero-shot & 0.077 & 13.40 & 14.59 \\
      & one-shot  & 0.074 & 13.29 & 14.42 \\
      & few-shot & 0.079 & 13.64 & 14.78 \\
      \bottomrule
      \end{tabular}
  \begin{tablenotes}[flushleft]
  \footnotesize
  \item NF (normalizing flow), NQ (NHITS+QRA), MO (Moirai), CX (ChronosX). Zero-shot includes no samples from DE-LU. One-shot contains one sample, while Few-shot 23 samples. Lower scores are better.
  \end{tablenotes}
  \end{threeparttable}
\end{table}

\clearpage

\section{Default Architecture of Proposed Models}
\label{sec:default_configs}
Tables~\ref{tab:e1_nf_defaults_all} and~\ref{tab:e1_nq_defaults_all} report the untuned default configurations used in E1 for the three size variants of each proposed model.

\newcolumntype{L}[1]{>{\raggedright\arraybackslash}p{#1}}
\newcolumntype{C}[1]{>{\centering\arraybackslash}p{#1}}


\setlength{\stripsep}{5pt}
\begin{strip}
\centering
\setlength{\tabcolsep}{4pt}
\renewcommand{\arraystretch}{1.05}
\captionof{table}{E1 NF default configuration}
\label{tab:e1_nf_defaults_all}
\scriptsize

\makebox[\textwidth][c]{%
\begin{tabular*}{\textwidth}{@{\extracolsep{\fill}} L{0.34\textwidth} C{0.22\textwidth} C{0.22\textwidth} C{0.22\textwidth}}
\toprule
\textbf{Parameter} & \textbf{tiny} & \textbf{small} & \textbf{base} \\
\midrule

\multicolumn{4}{l}{\textbf{Training configuration (common across sizes)}} \\
\texttt{n\_epochs}               & \multicolumn{3}{c}{\texttt{100}} \\
\texttt{batch\_size}             & \multicolumn{3}{c}{\texttt{128}} \\
\texttt{gradient\_clip\_val}     & \multicolumn{3}{c}{\texttt{2.0}} \\
\texttt{lr}                      & \multicolumn{3}{c}{\texttt{1e-4}} \\
\texttt{warmup\_epochs}          & \multicolumn{3}{c}{\texttt{2}} \\
\texttt{loss\_metric}            & \multicolumn{3}{c}{\texttt{es}} \\
\addlinespace[0.35em]

\multicolumn{4}{l}{\textbf{Model hyperparameters}} \\
\texttt{tf\_in\_size}     & \texttt{16}  & \texttt{96}  & \texttt{192} \\
\texttt{nf\_hidden\_dim}  & \texttt{16}  & \texttt{8}   & \texttt{192} \\
\texttt{n\_layers}        & \texttt{4}   & \texttt{10}  & \texttt{14} \\
\texttt{n\_heads}         & \texttt{2}   & \texttt{8}   & \texttt{8} \\
\texttt{n\_flow\_layers}  & \texttt{6}   & \texttt{10}  & \texttt{16} \\
\texttt{n\_made\_blocks}  & \texttt{1}   & \texttt{2}   & \texttt{2} \\
\texttt{tf\_dropout}      & \texttt{0.1} & \texttt{0.1} & \texttt{0.1} \\
\midrule
\# Weights & 49.0 K & 2.7 M & 19.0 M \\
\bottomrule
\end{tabular*}%
}
\end{strip}


\begin{strip}
\centering
\setlength{\tabcolsep}{4pt}
\renewcommand{\arraystretch}{1.05}
\captionof{table}{E1 NHITS+QRA default configuration}
\label{tab:e1_nq_defaults_all}
\scriptsize

\makebox[\textwidth][c]{%
\begin{tabular*}{\textwidth}{@{\extracolsep{\fill}} L{0.36\textwidth} C{0.213\textwidth} C{0.213\textwidth} C{0.213\textwidth}}
\toprule
\textbf{Parameter} & \textbf{tiny} & \textbf{small} & \textbf{base} \\
\midrule

\multicolumn{4}{l}{\textbf{NHITS training (common across sizes)}} \\
\texttt{n\_epochs}            & \multicolumn{3}{c}{\texttt{100}} \\
\texttt{batch\_size}          & \multicolumn{3}{c}{\texttt{128}} \\
\texttt{gradient\_clip\_val}  & \multicolumn{3}{c}{\texttt{1.0}} \\
\texttt{lr}                   & \multicolumn{3}{c}{\texttt{1e-3}} \\
\texttt{warmup\_epochs}       & \multicolumn{3}{c}{\texttt{2}} \\
\texttt{loss}                 & \multicolumn{3}{c}{\texttt{mae}} \\
\addlinespace[0.35em]

\multicolumn{4}{l}{\textbf{NHITS model}} \\
\texttt{n\_blocks} &
\texttt{[2,2]} &
\texttt{[2,2,2]} &
\texttt{[3,3,3]} \\
\texttt{mlp\_units} &
\texttt{[[16,16],[16,16]]} &
\texttt{[[96,96],[96,96],[96,96]]} &
\begin{tabular}[t]{@{}c@{}}
\texttt{[[256,256,256],}\\
\texttt{[256,256,256],}\\
\texttt{[256,256,256],}\\
\texttt{[256,256,256]]}
\end{tabular} \\
\texttt{dropout\_prob\_theta} & \texttt{0.1} & \texttt{0.1} & \texttt{0.1} \\
\texttt{n\_pool\_kernel\_size} & \texttt{[4,2]} & \texttt{[4,2,1]} & \texttt{[4,2,1]} \\
\texttt{n\_freq\_downsample}   & \texttt{[4,2]} & \texttt{[4,2,1]} & \texttt{[4,2,1]} \\
\addlinespace[0.35em]

\multicolumn{4}{l}{\textbf{QRA}} \\
\texttt{solver\_loss} & \texttt{iterative\_pinball} & \texttt{iterative\_pinball} & \texttt{iterative\_pinball} \\
\texttt{mc\_nhits\_samples\_qra} & \texttt{64} & \texttt{128} & \texttt{512} \\
\texttt{n\_epochs} & \texttt{200} & \texttt{200} & \texttt{200} \\
\texttt{batch\_size} & \texttt{512} & \texttt{512} & \texttt{1024} \\
\texttt{lr} & \texttt{1e-4} & \texttt{1e-4} & \texttt{1e-4} \\
\texttt{patience} & \texttt{10} & \texttt{20} & \texttt{20} \\
\texttt{use\_pca} & \texttt{true} & \texttt{true} & \texttt{true} \\
\texttt{pca\_var} & \texttt{0.95} & \texttt{0.95} & \texttt{0.95} \\
\texttt{sample\_k} & \texttt{0} & \texttt{0} & \texttt{0} \\
\texttt{lambda\_grid} &
\texttt{[0.0,1.0e-4,1.0e-3]} &
\texttt{[0.0,1.0e-4,1.0e-3]} &
\texttt{[0.0,1.0e-4,1.0e-3]} \\
\texttt{quantiles} &
\texttt{[1,10,50,90,99]} &
\texttt{[1,3,5,10,50,90,95,97,99]} &
\texttt{[1,3,5,10,50,90,95,97,99]} \\
\texttt{enforce\_monotonicity} & \texttt{true} & \texttt{true} & \texttt{true} \\
\texttt{repair\_method} & \texttt{isotonic} & \texttt{isotonic} & \texttt{isotonic} \\
\texttt{interp\_method} & \texttt{linear} & \texttt{linear} & \texttt{linear} \\
\texttt{target\_quantiles.uniform} & \texttt{200} & \texttt{200} & \texttt{200} \\
\texttt{subsample\_stride} & \texttt{4} & \texttt{2} & \texttt{1} \\
\addlinespace[0.35em]
\multicolumn{4}{l}{\textbf{SWAG (common across sizes)}} \\
\texttt{swag.enabled}         & \multicolumn{3}{c}{\texttt{true}} \\
\texttt{swag.start\_epoch}     & \multicolumn{3}{c}{\texttt{5}} \\
\texttt{swag.collect\_every}   & \multicolumn{3}{c}{\texttt{1}} \\
\texttt{swag.max\_rank}        & \multicolumn{3}{c}{\texttt{20}} \\
\texttt{swag.var\_clamp}       & \multicolumn{3}{c}{\texttt{1e-30}} \\
\texttt{swag.scale}            & \multicolumn{3}{c}{\texttt{1.0}} \\
\midrule
\# Weights & 88.7 K & 1.5 M & 6.3 M \\
\bottomrule
\end{tabular*}%
}
\end{strip}

\FloatBarrier
\clearpage
\section{Hyperparameter-Tuning}
\label{sec:hyptun}

\newcolumntype{L}[1]{>{\raggedright\arraybackslash}p{#1}}
\newcolumntype{C}[1]{>{\centering\arraybackslash}p{#1}}

\begin{strip}
\centering
\setlength{\tabcolsep}{4pt}
\renewcommand{\arraystretch}{1.05}
\captionof{table}{E2 NF tuned configuration}
\label{tab:e2_nf_tuned_all}
\scriptsize

\makebox[\textwidth][c]{%
\begin{tabular*}{\textwidth}{@{\extracolsep{\fill}} L{0.34\textwidth} C{0.22\textwidth} C{0.22\textwidth} C{0.22\textwidth}}
\toprule
\textbf{Parameter} & \textbf{tiny} & \textbf{small} & \textbf{base} \\
\midrule

\multicolumn{4}{l}{\textbf{Training configuration (common across sizes)}} \\
\texttt{n\_epochs}               & \multicolumn{3}{c}{\texttt{100}} \\
\texttt{batch\_size}             & \multicolumn{3}{c}{\texttt{128}} \\
\texttt{gradient\_clip\_val}     & \multicolumn{3}{c}{\texttt{2.0}} \\
\texttt{lr}                      & \texttt{4.438e-5} & \texttt{1.7928e-4} & \texttt{3.7903e-5} \\
\texttt{warmup\_epochs}          & \texttt{3} & \texttt{1} & \texttt{1} \\
\texttt{loss\_metric}            & \multicolumn{3}{c}{\texttt{es}} \\
\addlinespace[0.35em]

\multicolumn{4}{l}{\textbf{Model hyperparameters}} \\
\texttt{tf\_in\_size}     & \texttt{16}  & \texttt{96}  & \texttt{256} \\
\texttt{nf\_hidden\_dim}  & \texttt{16}  & \texttt{128} & \texttt{256} \\
\texttt{n\_layers}        & \texttt{6}   & \texttt{10}  & \texttt{16} \\
\texttt{n\_heads}         & \texttt{4}   & \texttt{8}   & \texttt{4} \\
\texttt{n\_flow\_layers}  & \texttt{8}   & \texttt{12}  & \texttt{18} \\
\texttt{n\_made\_blocks}  & \texttt{1}   & \texttt{1}   & \texttt{2} \\
\texttt{tf\_dropout}      & \texttt{0.285}  & \texttt{0.229}  & \texttt{0.2678} \\
\midrule
\# Weights & 69.0 K & 3.5 M & 38.2 M \\
\bottomrule
\end{tabular*}%
}
\end{strip}


\begin{strip}
\centering
\setlength{\tabcolsep}{4pt}
\renewcommand{\arraystretch}{1.05}
\captionof{table}{E2 NHITS+QRA tuned configuration}
\label{tab:e2_nq_tuned_all}
\scriptsize

\makebox[\textwidth][c]{%
\begin{tabular*}{\textwidth}{@{\extracolsep{\fill}} L{0.36\textwidth} C{0.213\textwidth} C{0.213\textwidth} C{0.213\textwidth}}
\toprule
\textbf{Parameter} & \textbf{tiny} & \textbf{small} & \textbf{base} \\
\midrule

\multicolumn{4}{l}{\textbf{NHITS training (common across sizes)}} \\
\texttt{n\_epochs}            & \multicolumn{3}{c}{\texttt{100}} \\
\texttt{batch\_size}          & \multicolumn{3}{c}{\texttt{128}} \\
\texttt{gradient\_clip\_val}  & \multicolumn{3}{c}{\texttt{1.0}} \\
\texttt{lr}                   & \texttt{5.552e-5} & \texttt{1.1929e-4} & \texttt{6.1802e-5} \\
\texttt{warmup\_epochs}       & \multicolumn{3}{c}{\texttt{2}} \\
\texttt{loss}                 & \multicolumn{3}{c}{\texttt{mae}} \\
\addlinespace[0.35em]

\multicolumn{4}{l}{\textbf{NHITS model}} \\
\texttt{n\_blocks} &
\texttt{[1,1]} &
\texttt{[2,2,2,2]} &
\texttt{[2,2,2,2]} \\
\texttt{mlp\_units} &
\texttt{[[32,32],[32,32]]} &
\begin{tabular}[t]{@{}c@{}}
\texttt{[[96,96],[96,96],}\\
\texttt{[96,96],[96,96]]}
\end{tabular}
&
\begin{tabular}[t]{@{}c@{}}
\texttt{[[256,256,256],} \\
\texttt{[256,256,256],}\\
\texttt{[256,256,256],} \\
\text{[256,256,256]]}
\end{tabular}
\\
\texttt{dropout\_prob\_theta} & \texttt{0.154} & \texttt{0.141} & \texttt{0.1183} \\
\texttt{n\_pool\_kernel\_size} & \texttt{[2,2]} & \texttt{[8,4,2,2]} & \texttt{[16,8,4,2]} \\
\texttt{n\_freq\_downsample}   & \texttt{[2,2]} & \texttt{[4,2,2,1]} & \texttt{[16,8,4,2]} \\
\addlinespace[0.35em]

\multicolumn{4}{l}{\textbf{QRA}} \\
\texttt{solver\_loss} & \texttt{iterative\_pinball} & \texttt{iterative\_pinball} & \texttt{iterative\_pinball} \\
\texttt{mc\_nhits\_samples\_qra} & \texttt{8} & \texttt{48} & \texttt{128} \\
\texttt{n\_epochs} & \texttt{200} & \texttt{200} & \texttt{100} \\
\texttt{batch\_size} & \texttt{512} & \texttt{1024} & \texttt{256} \\
\texttt{lr} & \texttt{1.02e-4} & \texttt{1.6597e-4} & \texttt{9.16241e-5} \\
\texttt{patience} & \texttt{10} & \texttt{20} & \texttt{30} \\
\texttt{use\_pca} & \texttt{false} & \texttt{false} & \texttt{false} \\
\texttt{sample\_k} & \texttt{1} & \texttt{1} & \texttt{1} \\
\texttt{lambda\_grid} &
\texttt{[0.0, 1.0e-3]} &
\texttt{[0.0, 1.0e-3]} &
\texttt{[0.0, 1.0e-4, 1.0e-3]} \\
\texttt{quantiles} &
\texttt{[1,10,50,90,99]} &
\texttt{[1,3,5,10,50,90,95,97,99]} &
\begin{tabular}[t]{@{}c@{}}
\texttt{[1,3,5,10,50,90,}\\
\texttt{95,97,99]}
\end{tabular}
\\
\texttt{enforce\_monotonicity} & \texttt{true} & \texttt{true} & \texttt{true} \\
\texttt{repair\_method} & \texttt{isotonic} & \texttt{isotonic} & \texttt{isotonic} \\
\texttt{interp\_method} & \texttt{linear} & \texttt{linear} & \texttt{linear} \\
\texttt{target\_quantiles.uniform} & \texttt{200} & \texttt{200} & \texttt{200} \\
\texttt{subsample\_stride} & \texttt{2} & \texttt{2} & \texttt{1} \\
\addlinespace[0.35em]

\multicolumn{4}{l}{\textbf{SWAG (only where enabled)}} \\
\texttt{swag.enabled}       & \texttt{false} & \texttt{true} & \texttt{true} \\
\texttt{swag.start\_epoch}  &  & \texttt{5} & \texttt{5} \\
\texttt{swag.collect\_every}&  & \texttt{4} & \texttt{4} \\
\texttt{swag.max\_rank}     &  & \texttt{10} & \texttt{10} \\
\texttt{swag.var\_clamp}    &  & \texttt{1e-30} & \texttt{1e-30} \\
\texttt{swag.scale}         &  & \texttt{0.5} & \texttt{0.5} \\
\midrule
\# Weights & 115.3 K & 1.2 M & 3.9 M \\
\bottomrule
\end{tabular*}%
}
\end{strip}

\FloatBarrier
\clearpage
\section{Carbon Footprinting}
\label{sec:carbon}

Emissions are estimated using a carbon intensity of 328\,gCO\textsubscript{2}e/kWh (2025 German average~\cite{CO2EmissionsKWh}) and a power usage effectiveness (PUE) of 1.2.

\begin{strip}
\centering
\setlength{\tabcolsep}{6pt}
\renewcommand{\arraystretch}{1.15}
\captionof{table}{Energy and CO\textsubscript{2}e summary of E1 by model, size, and phase.}
\label{tab:e1_energy_co2}
\begin{tabular}{@{} l l l r r r r @{}}
\toprule
Model & Size & Phase & Time (h) & Energy (kWh) & CO\textsubscript{2}e (kg) & CO\textsubscript{2}e$\cdot$PUE (kg) \\
\midrule

\multirow{6}{*}{NF}
  & \multirow{2}{*}{tiny}  & train & 0.292 & 0.30 & 0.10 & 0.12 \\
  &                        & test  & 0.011 & 0.01 & 0.00 & 0.00 \\
  & \multirow{2}{*}{small} & train &  0.789 & 1.07	& 0.35	& 0.42     \\
  &                        & test  & 0.030 & 0.02 & 0.01 & 0.01 \\
  & \multirow{2}{*}{base}  & train & 1.182 & 1.11 & 0.36 & 0.44 \\
  &                        & test  & 0.022 & 0.01 & 0.00 & 0.01 \\
\addlinespace[0.35em]

\multirow{6}{*}{NQ}
  & \multirow{2}{*}{tiny}  & train & 0.938 & 0.89 & 0.29 & 0.35 \\
  &                        & test  & 0.038 & 0.03 & 0.01 & 0.01 \\
  & \multirow{2}{*}{small} & train & 1.674 & 1.34 & 0.44 & 0.53 \\
  &                        & test  & 0.225 & 0.21 & 0.07 & 0.08 \\
  & \multirow{2}{*}{base}  & train & 3.196 & 4.79 & 1.57 & 1.89 \\
  &                        & test  & 0.422 & 0.57 & 0.19 & 0.23 \\
\addlinespace[0.35em]

\multirow{3}{*}{MO-ZS}
  & small & test & 0.022 & 0.01 & 0.00 & 0.01 \\
  & base  & test & 0.028 & 0.02 & 0.01 & 0.01 \\
  & large & test & 0.012 & 0.01 & 0.00 & 0.00 \\
\addlinespace[0.35em]

\multirow{6}{*}{MO}
  & \multirow{2}{*}{small} & train & 0.243 & 0.36 & 0.12 & 0.14 \\
  &                        & test  & 0.011 & 0.01 & 0.00 & 0.00 \\
  & \multirow{2}{*}{base}  & train & 0.199 & 0.37 & 0.12 & 0.14 \\
  &                        & test  & 0.022 & 0.01 & 0.00 & 0.01 \\
  & \multirow{2}{*}{large} & train & 0.398 & 0.67 & 0.22 & 0.26 \\
  &                        & test  & 0.013 & 0.01 & 0.00 & 0.00 \\
\addlinespace[0.35em]

\multirow{3}{*}{CX-ZS}
  & small & test & 0.035 & 0.03 & 0.01 & 0.01 \\
  & base  & test & 0.036 & 0.04 & 0.01 & 0.02 \\
  & large & test & 0.097 & 0.07 & 0.02 & 0.03 \\
\addlinespace[0.35em]

\multirow{6}{*}{CX}
  & \multirow{2}{*}{small} & train & 0.467 & 0.49 & 0.16 & 0.19 \\
  &                        & test  & 0.029 & 0.01 & 0.00 & 0.00 \\
  & \multirow{2}{*}{base}  & train & 0.942 & 1.41 & 0.46 & 0.55 \\
  &                        & test  & 0.034 & 0.02 & 0.01 & 0.01 \\
  & \multirow{2}{*}{large} & train & 2.871 & 3.54 & 1.16 & 1.39 \\
  &                        & test  & 0.055 & 0.05 & 0.02 & 0.02 \\
\bottomrule
\end{tabular}
\end{strip}




\clearpage
\setlength{\stripsep}{5pt}
\begin{strip}
\centering
\setlength{\tabcolsep}{6pt}
\renewcommand{\arraystretch}{1.15}
\captionof{table}{E3 Energy and CO\textsubscript{2}e summary for feature evaluation (training) and selected feature groups (test)}
\label{tab:energy_co2_feature_eval}
\begin{tabular}{@{} l l l r r r r @{}}
\toprule
Model--Size & Phase & Group & Time (h) & Energy (kWh) & CO\textsubscript{2}e (kg) & CO\textsubscript{2}e$\cdot$PUE (kg) \\
\midrule

\multicolumn{7}{@{}l}{\textbf{NF-small}} \\
\addlinespace[0.15em]
Feature evaluation (train) & train &  & 2.333 & 2.03 & 0.67 & 0.80 \\
\addlinespace[0.45em]

\multicolumn{7}{@{}l}{\textbf{NQ-tiny}} \\
\addlinespace[0.15em]
Feature evaluation (train) & train &  & 4.370 & 5.62 & 1.84 & 2.21 \\
Selected features (test)   & test  & \texttt{R3} & 0.020 & 0.02 & 0.01 & 0.01 \\
Selected features (test)   & test  & \texttt{R4} & 0.030 & 0.15 & 0.05 & 0.06 \\
Selected features (test)   & test  & \texttt{R5} & 0.019 & 0.02 & 0.01 & 0.01 \\
\addlinespace[0.45em]

\multicolumn{7}{@{}l}{\textbf{MO-base}} \\
\addlinespace[0.15em]
Feature evaluation (train) & train &  & 8.301 & 10.33 & 3.38 & 4.06 \\
Selected features (test)   & test  & \texttt{R1} & 0.023 & 0.01 & 0.00 & 0.01 \\
Selected features (test)   & test  & \texttt{R2} & 0.022 & 0.02 & 0.01 & 0.01 \\
Selected features (test)   & test  & \texttt{R3} & 0.010 & 0.01 & 0.00 & 0.00 \\
\addlinespace[0.45em]
\multicolumn{7}{@{}l}{\textbf{CX-small}} \\
\addlinespace[0.15em]
Feature evaluation (train) & train &  & 2.073 & 1.69 & 0.55 & 0.66 \\
\bottomrule
\end{tabular}
\end{strip}

\begin{strip}
\centering
\setlength{\tabcolsep}{6pt}
\renewcommand{\arraystretch}{1.15}
\captionof{table}{E4 Energy and CO\textsubscript{2}e summary by model/size, learning strategy, and phase.}
\label{tab:energy_co2_strategy_nf_nq_mo}
\begin{tabular}{@{} l l l r r r r @{}}
\toprule
Model--Size & Learning Strategy & Phase & Time (h) & Energy (kWh) & CO\textsubscript{2}e (kg) & CO\textsubscript{2}e$\cdot$PUE (kg) \\
\midrule

\multirow{6}{*}{NF--small}
  & \multirow{2}{*}{Zero-shot} & train & 3.642   & 2.76  & 0.91 & 1.09 \\
  &                            & test  & 0.026 & 0.02  & 0.01 & 0.01 \\ 
  & \multirow{2}{*}{One-shot}  & train & 4.190   & 3.01  & 0.99 & 1.18 \\
  &                            & test  & 0.014 & 0.01  & 0.00 & 0.00 \\ 
  & \multirow{2}{*}{Few-shot}  & train & 13.002   & 10.10 & 3.31 & 3.98 \\
  &                            & test  & 0.025 & 0.02  & 0.01 & 0.01 \\ 
\addlinespace[0.35em]

\multirow{6}{*}{NQ--tiny}
  & \multirow{2}{*}{Zero-shot} & train & 13.921   & 1.06 & 0.35 & 0.42 \\
  &                            & test  & 0.034 & 0.03 & 0.01 & 0.01 \\ 
  & \multirow{2}{*}{One-shot}  & train & 13.684   & 1.14 & 0.38 & 0.45 \\
  &                            & test  & 0.032 & 0.03 & 0.01 & 0.01 \\ 
  & \multirow{2}{*}{Few-shot}  & train & 14.033   & 1.21 & 0.40 & 0.48 \\
  &                            & test  & 0.019 & 0.01 & 0.00 & 0.01 \\ 
\addlinespace[0.35em]

\multirow{6}{*}{MO--base}
  & \multirow{2}{*}{Zero-shot} & train & 6.110 & 10.97 & 3.60 & 4.32 \\ 
  &                            & test  & 0.019 & 0.01  & 0.00 & 0.00 \\ 
  & \multirow{2}{*}{One-shot}  & train & 6.227 & 10.37 & 3.40 & 4.08 \\ 
  &                            & test  & 0.022 & 0.01  & 0.00 & 0.01 \\ 
  & \multirow{2}{*}{Few-shot}  & train & 6.234 & 10.41 & 3.41 & 4.10 \\ 
  &                            & test  & 0.024 & 0.01  & 0.00 & 0.00 \\ 
  
\addlinespace[0.35em]

\multirow{6}{*}{CX--small}
  & \multirow{2}{*}{Zero-shot} & train & 9.235 & 7.39 & 2.42 & 2.91 \\ 
  &                            & test  & 0.019 & 0.01  & 0.00 & 0.01 \\ 
  & \multirow{2}{*}{One-shot}  & train & 7.748 & 6.19 & 2.03 & 2.44 \\ 
  &                            & test  & 0.018 & 0.01  & 0.00 & 0.01 \\ 
  & \multirow{2}{*}{Few-shot}  & train & 8.37 & 6.92 & 2.27 & 2.71 \\ 
  &                            & test  & 0.018 & 0.02  & 0.01 & 0.01 \\ 
\bottomrule
\end{tabular}
\end{strip}

\clearpage
\printcredits

\section*{Declaration of generative AI and AI-assisted technologies in the manuscript preparation process}

During the preparation of this work, the authors used OpenAI ChatGPT (o3-mini/o3-mini-high, GPT-4o, and GPT-5) and Anthropic Claude (Opus 4.7 and 5) to support both experimental work and manuscript preparation. For the experimental work, these tools were used to improve code readability, assist with debugging, prototype user interfaces, enhance visualizations, and support the refinement of experimental methodologies. For manuscript preparation, they were used to assist with text summarization, structural coherence, table and \LaTeX{} formatting, and prose refinement.

After using these tools and services, the authors reviewed and edited the outputs as needed and take full responsibility for the content of the published article.

\bibliographystyle{cas-model2-names}

\bibliography{ref}



\end{document}